\UseRawInputEncoding
\documentclass[12pt,a4paper]{article}
\usepackage[utf8]{inputenc}
\usepackage{amsmath}
\usepackage{xcolor} 
\usepackage[normalem]{ulem}
\usepackage{setspace}
\usepackage{microtype}
\usepackage[margin=1in]{geometry}
\usepackage[utf8]{inputenc} 
\usepackage[T1]{fontenc}    
\usepackage{url}            
\usepackage{booktabs}       
\usepackage{amsfonts}       
\usepackage{nicefrac}       
\usepackage{graphicx}
\usepackage{float}
\usepackage{hyperref}       
\usepackage{makecell}

\title{\textbf{MatMMExtract: An Open-Source Pipeline for Panel-Level Extraction of Grounded Image-Text Pairs from Materials Science Literature}}

\author{
  Subham Ghosh$^{1}$, Shubham Tiwari$^{2}$,\\[2pt]
  Mohammad Ibrahim$^{1}$, and Abhishek Tewari$^{1,2,*}$ \\[15pt]
  $^{1}$Mehta Family School of Data Science and Artificial Intelligence\\Indian Institute of Technology Roorkee, Uttarakhand, India-247667\\[4pt]
  $^{2}$Department of Metallurgical and Materials Engineering\\Indian Institute of Technology Roorkee, Uttarakhand, India-247667 \\[10pt]
  $^{*}$\textbf{Corresponding author email:} \texttt{abhishek@mt.iitr.ac.in}
}

\makeatletter
\renewcommand\normalsize{%
   \@setfontsize\normalsize{12pt}{12pt} 
}
\makeatother
\normalsize

\begin{document}
\maketitle
\textbf{Keywords:} LLM \ensuremath{\cdot} Multimodal dataset \ensuremath{\cdot} Information extraction \ensuremath{\cdot} Compound figure detection \ensuremath{\cdot} Materials informatics 

\vspace{1em}

\begin{abstract}

The materials science literature encodes decades of experimental knowledge in figures, yet this visual record remains inaccessible to AI at scale. The core difficulty is structural: most scientific figures are compound, with a single caption describing multiple sub-panels simultaneously, making direct image-text pairing unreliable. We present \textbf{MatMMExtract}, an end-to-end open-source pipeline that resolves this by decomposing compound figures into individual sub-panels and generating structured, grounded annotations using a large language model guided by a curated materials science taxonomy. Applied to 14,810 open-access articles, MatMMExtract produces \textbf{MatSciFig} - 391,606 panel-level image-text pairs from 180,571 figures, each annotated with a sub-caption, a two-level visualisation category spanning 19 classes and over 100 subtypes, and a scientific summary. To enable accurate panel localisation, we introduce \textbf{MaterialScope}, a domain-specific detection dataset of 2,811 manually annotated materials science figures, on which a fine-tuned YOLO12-m detector achieves mAP$_{50}$ of 0.9227. Among six benchmarked language models, Gemini 3.1 Flash Lite delivers the best cost-quality trade-off for annotation generation, with 80.3\% of sub-captions and 81.8\% of summaries rated good and a hallucination rate of 5.4\%. A dual-encoder retrieval baseline on MatSciFig achieves a $4.7\times$ improvement in R@1 over zero-shot CLIP, demonstrating the dataset's immediate utility for vision-language learning. This work lays the groundwork for extending reasoning-oriented multimodal supervision to materials science. All resources are released openly to the community.
\end{abstract}

\section{Introduction}
Materials discovery is increasingly driven by data and machine learning, which now includes property prediction, inverse design, and the analysis of
structure-property relationships~\cite{PEIVASTE2025119419, SHENGCAO202691}. The community has built high-throughput repositories such as the Materials Project~\cite{jain2013commentary}, the OQMD~\cite{saal2013materials}, and the Materials Data Facility~\cite{blaiszik2019data}. Yet such structured databases capture only a narrow slice of materials knowledge, most of them recording tabulated property values while omitting the experimental context, processing history, and qualitative interpretation that authors encode in the prose and figures of their articles. More importantly, multimodal datasets present significant challenges in the scientific domain, since most data remains fossilised within research articles in forms that resist automated consumption~\cite{gottweis2025towards,khalighinejad-etal-2025-matvix}. While significant progress has been made in constructing image-text large datasets for the medical domain~\cite{johnson2019mimic,bannur2023learning,ruckert2024rocov2,baghbanzadeh2025open,lozano2025biomedica,song-etal-2025-figex} and some for interdisciplinary scientific domains~\cite{li2024mmsci,li-etal-2024-multimodal-arxiv,tao2026omniscience}, the materials science domain remains largely underexplored, with no large-scale multimodal dataset capturing the diverse visual modalities inherent to the field. 

Materials science research is inherently multi-modal, encompassing a wide variety of image types, including microscopy, diffraction, spectroscopy, generic plots, and many more. Embedded within millions of published articles, these figures contain rich experimental data describing material properties, synthesis conditions, and characterisation results, yet they remain largely inaccessible to modern AI pipelines. Recent advances in vision-language models such as CLIP~\cite{radford2021learning}, BLIP~\cite{li2023blip}, LLaVA~\cite{liu2023visual}, and many others have demonstrated remarkable capabilities for aligning visual and textual representations across diverse domains. However, these models struggle with domain-specific scientific imagery, largely due to a lack of high-quality, domain-aligned training data.

Existing efforts in materials science to collect multimodal data remain limited in scope. Exsclaim~\cite{schwenker2023exsclaim} automates figure extraction from materials science literature but remains narrowly scoped to microscopy with subfigure labels A–H. More recent resources, such as MatCha~\cite{lai-etal-2025-multimodal}, MatQnA~\cite{weng2025matqna}, and MATRIX~\cite{mcgrath2026matrix}, provide expert-level evaluation benchmarks, but none deliver the large-scale, modality-diverse multimodal training corpus that the field critically lacks. MatMech~\cite{liu2026multimodal}, while large-scale, focuses exclusively on causal reasoning chains rather than the diverse image-text pairs needed for vision-language training across the full spectrum of materials characterisation modalities. Constructing such a corpus is non-trivial, as the majority of published figures are compound~\cite{taschwer2018automatic}, and object detectors for decomposition require domain-specific annotated data.  

Compound figure detection has gone through many technical generations. It began with rule-based methods from the ImageCLEF challenges, using whitespace and grid layouts~\cite{de2016overview}; later, more robust CNNs improved results~\cite{tsutsui2017data,zou2020unified}. In Exsclaim~\cite{schwenker2023exsclaim}, researchers use YOLOv3~\cite{redmon2018yolov3} along with ResNet-152~\cite{he2016deep} for layout and labels. Recent YOLO models~\cite{yolov8_ultralytics,wang2024yolov9,wang2024yolov10,yolo11_ultralytics,tian2025yolov12} and detection transformers~\cite{liu2022dabdetr,song-etal-2025-figex} are capable of performing both tasks. Prior work shows how to use LLMs for scientific caption generation~\cite{hsu2026large,hsu2023gpt,timklaypachara2025leveraging}. Prompting matters as much as the model for this task. FigEx~\cite{song-etal-2025-figex} and OmniScience~\cite{tao2026omniscience} show that scoping the prompt and right context actually improves generation. 

Experience from the biomedical domain tells that treating compound figures as atomic units introduces noise that degrades learned representations~\cite{baghbanzadeh2025open}, and that raw figure captions provide weak image-text alignment without semantic enrichment from surrounding text~\cite{li-etal-2024-multimodal-arxiv}. Even once panels are localised, caption-only supervision is fundamentally impoverished, providing no grounding of each panel in its surrounding scientific argument and no consistent taxonomy of the heterogeneous visualisation types encountered in materials research. Moreover, the absence of panel-level summaries of scientific significance means such supervision cannot support the rich, context-aware understanding that materials science vision-language models demand.

In this work, we address these deficits with \textbf{MatMMExtract}, an open-source pipeline that transforms open-access materials science articles into a panel-level multimodal dataset, and we use it to release \textbf{MatSciFig}, the resulting large-scale resource. Our contributions are as follows:

\begin{itemize}
    \item \textbf{MatMMExtract pipeline.} An open-source, license-aware pipeline that parses publisher XML into $\langle$\emph{image, caption, reference}$\rangle$ triplets and decomposes compound figures into panels with grounded sub-captions, category/subtype labels, and summaries; released as the \texttt{matmmextract} PYPI package.

    \item \textbf{MaterialScope benchmark.} $2{,}811$ manually annotated figures ($12{,}906$ boxes) across diverse materials domains; we benchmark six detectors and show domain-specific training drives accuracy.

    \item \textbf{Taxonomy and annotation protocol.} A two-level taxonomy of $19$ visualisation categories, with six low-cost LLMs benchmarked on classification, caption/summary quality, and hallucination to identify reliable annotators.

    \item \textbf{MatSciFig dataset.} $391{,}606$ annotated panels from $180{,}571$ figures, one of the largest figure-level vision language datasets in materials science. Each panel carries a sub-caption, a two-level visualisation category and subtype label, and a scientific summary.

    \item \textbf{Downstream utility.} A CLIP-style dual-encoder combining CLIP ViT-B/32 and MatSciBERT trained on MatSciFig (indexed with FAISS~\cite{8733051}), providing a reference point for future vision-language research on scientific imagery.
\end{itemize}

\section{Methods}
\begin{figure}[H]
    \centering
    \includegraphics[width=0.99\linewidth]{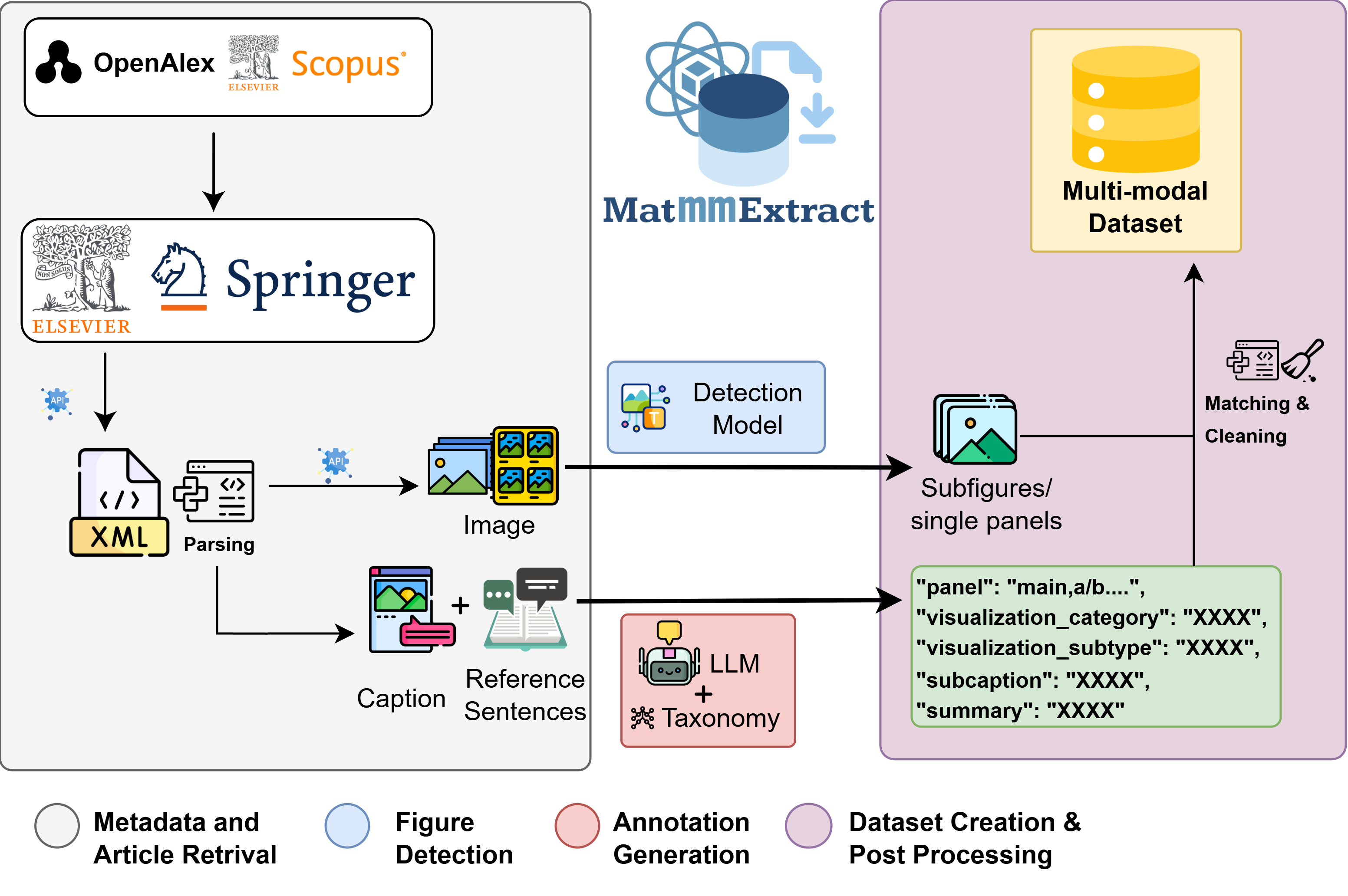}
    \caption{Overview of the MatMMExtract pipeline. Figures, captions, and reference sentences are extracted from Elsevier and Springer articles, decomposed into sub-panels by a compound figure detector, and annotated using an LLM guided by a materials science taxonomy to produce the final multimodal dataset.}
    \label{fig:pipeline}
\end{figure}

Figure~\ref{fig:pipeline} provides an overview of the MatMMExtract pipeline, which consists of three sequential stages. First, full-text articles are retrieved from Elsevier and Springer publishers, and figures, captions, and in-text reference sentences are extracted from the source XML. Second, a compound figure detection model identifies and localises individual sub-panels within each figure. Third, an LLM conditioned on the caption, reference sentence, and a predefined materials science taxonomy generates structured annotations for each sub-panel. And finally, combining the output of steps two and three creates the final multimodal dataset. The following subsections describe each stage in detail.
\subsection{Source Data Collection}
We curate materials science literature with a pipeline that begins by collecting metadata from two complementary databases, OpenAlex~\cite{priem2022openalex} and Scopus. OpenAlex supports fine-grained license filtering, so we can select articles by specific Creative Commons type, such as CC-BY, CC-BY-NC, and CC-BY-NC-ND. Scopus is coarser: it labels open-access articles only as gold or hybrid, categories that may include licenses that restrict derivative works. To ensure compatibility with downstream datasets and their redistribution, we restrict our corpus to \textbf{CC-BY} and \textbf{CC-BY-NC} licensed articles. Following metadata collection, full-text content is retrieved through publisher APIs. For \textbf{Elsevier} and \textbf{Springer} journals, we obtain structured \textbf{XML} files\footnote{TDM agreements govern bulk content scraping and downloading, distinct from standard library subscriptions, and are required to avoid impacting publisher server performance.}. We parse these XML files to extract three key components: figure image links, their corresponding captions, and the in-text reference sentences where each figure is cited within the body of the article. The figure images are subsequently downloaded using the extracted links. As graphical abstracts have no captions, we remove them. Following this filtering step, each remaining entry forms a structured \textbf{triplet} of 
$\langle$\textit{image, caption, in-text reference}$\rangle$, providing richer context than caption-only pairs.

A significant challenge in processing these figures is the prevalence of compound figures. Based on open-access biomedical literature, approximately 40 - 60\% of figures in published articles are compound figures~\cite{taschwer2018automatic}. Consistent with this observation, we find that $62\%$ of figures in our curated corpus are compound figures, posing a challenge for direct image-text alignment since a single caption describes multiple sub-images simultaneously. To address this, we decompose the processing pipeline into two sequential stages. In the first stage, we classify each figure as either a single image or a compound figure, containing multiple sub-panels labelled (a),(b),(c),(d), etc., while simultaneously converting them into atomic ones. In the second stage, we leverage both the figure caption and the in-text reference sentence to generate structured annotations for each sub-panel. Specifically, for each atomic image unit, we generate (i) a \textbf{caption}, grounded from the relevant portion of the original figure caption and in-text reference sentence; (ii) a \textbf{visualisation category and sub-category}, classifying the image type (e.g., microscopy $\rightarrow$ SEM, TEM, diffraction $\rightarrow$ XRD, EBSD Map, etc); and (iii) a concise \textbf{summary} of the image content and its scientific significance.

\subsection{Compound Figure Detection}
\label{sec:compound_figure_detection}
A key requirement of our pipeline is to reliably detect whether a figure is a single image or a compound figure, and to localise individual sub-panels when present. To identify the best-performing object detector for this task, we conducted a systematic benchmark across a range of modern architectures. We initially leveraged annotations from FigEx~\cite{song-etal-2025-figex}, a tool developed for scientific figure segmentation, though primarily validated on biomedical literature. Using this data as a starting point (6,454 train / 717 val images), we trained the following detectors from pretrained weights for 50 epochs: YOLO8-m~\cite{yolov8_ultralytics}, YOLO9-c~\cite{wang2024yolov9}, YOLO10-m~\cite{wang2024yolov10}, YOLO11-m~\cite{yolo11_ultralytics}, YOLO12-m~\cite{tian2025yolov12}, and DAB-DETR~\cite{liu2022dabdetr}. Across all the models, average precision on the frequent class (APf) ranged from 0.67 to 0.73 on  MaterialScope (curated materials dataset), which was notably lower than the APf values (>0.9) achieved on the FigEx dataset, further confirming the domain gap between biomedical and materials imagery (full results in Table~\ref{tab:figex_val}, and Table~\ref{tab:figex_baseline}, Appendix~\ref{sec:appendix_figex_baseline}).
\subsubsection{Curated Dataset}
\begin{figure}[H]
    \centering
    \includegraphics[width=0.99\linewidth]{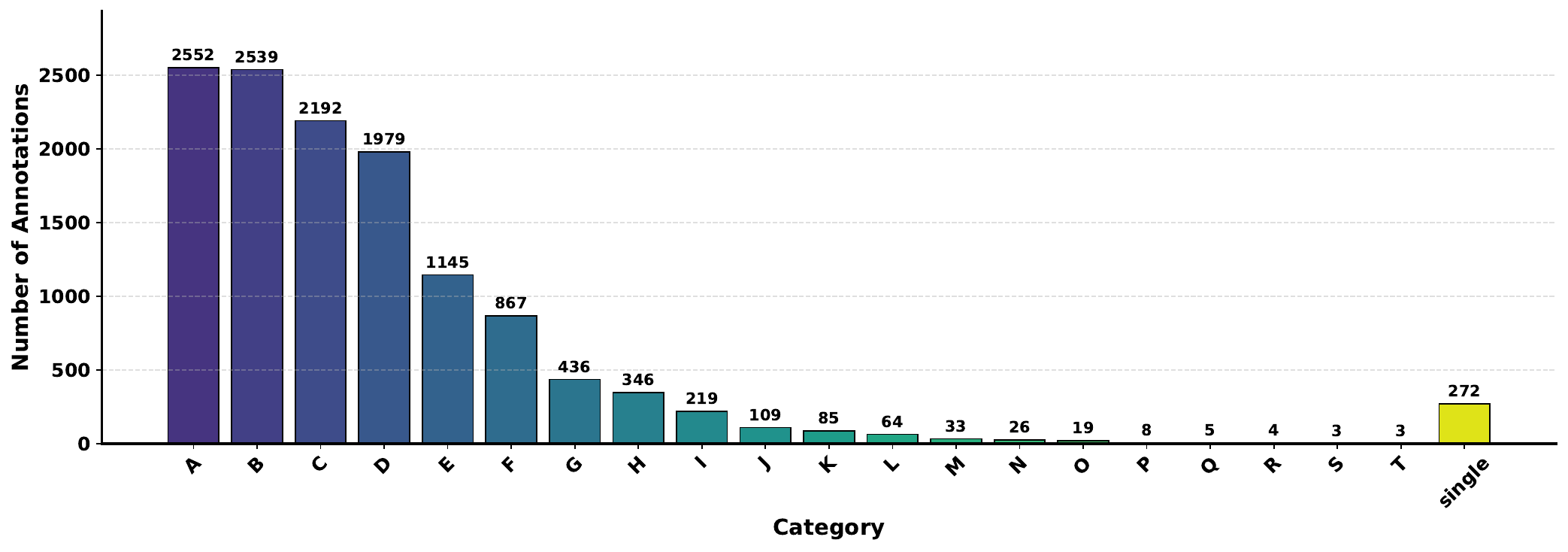}
    \caption{Distribution of valid bounding box annotations across sub-panel labels}
    \label{fig:annotation_dist}
\end{figure}

To address the domain gap, we curated a materials-science-specific annotation dataset. A total of 2,811 figures were manually annotated using Label Studio, drawn from various materials science domains including alloys, composites, ceramics, steels, polymers, and thin films. All annotations were carried out by two PhD-level researchers with domain expertise, who independently labelled sub-panel bounding boxes for each figure, yielding 12,906 valid bounding box annotations across sub-panel labels A-T and single-panel figures. The distribution of annotations across sub-panel labels is summarised in Figure~\ref{fig:annotation_dist}. We named this dataset MaterialScope.

To assess the reliability of the annotation process, both annotators labelled a shared subset of 102 images, and inter-annotator agreement was computed using Cohen's Kappa over the complete panel-label set assigned to each figure, where two annotations are considered to agree if and only if they produce identical sets of panel letters. The resulting score of $\kappa = 0.9245$ indicates near-perfect agreement~\cite{Landis1977TheMO}, lending strong confidence to the quality and consistency of the curated annotations. 
The MaterialScope is partitioned into training (2,249), validation (281), and test (281) sets, stratified to preserve the sub-panel label distribution across all three subsets.

\subsubsection{Domain-Adaptive Training}
\label{sec:domain_adaptive_training}
Using the MaterialScope, we retrain the same six architectures, YOLO8-m~\cite{yolov8_ultralytics}, YOLO9-c~\cite{wang2024yolov9}, YOLO10-m~\cite{wang2024yolov10}, YOLO11-m~\cite{yolo11_ultralytics}, YOLO12-m~\cite{tian2025yolov12}, and DAB-DETR~\cite{liu2022dabdetr}, to directly quantify the benefit of domain-specific annotation. All YOLO-family models are fine-tuned from official pre-trained weights for 50 epochs at $1024 \times 1024$ resolution with a batch size of 16, using the Ultralytics default SGD optimiser with momentum 0.937 and cosine learning-rate decay. DAB-DETR~\cite{liu2022dabdetr} is trained for 50 epochs with a batch size of 8 and gradient accumulation of 2 using AdamW (backbone lr\,=\,$10^{-5}$, head lr\,=\,$10^{-4}$, weight decay\,=\,$10^{-4}$), with a 2-epoch linear warmup followed by cosine decay and early stopping with patience 5. All experiments are conducted on a single NVIDIA RTX A6000 48\,GB GPU.

\subsubsection{Experimental Setup}
All models are evaluated on the held-out test split under a unified protocol: confidence threshold 0.55 (for precision, recall, and F1), NMS IoU threshold 0.45, and inference resolution $1024 \times 1024$. Precision, recall, and F1 are computed via greedy IoU matching at an IoU threshold of 0.50. We also report frequency-stratified average precision: APf denotes mean AP@50:95 on frequent sub-panel classes (A-E, ${>}100$ test panels), and APc on common classes (F-K and \textit{single}, 10 to 100 panels). Throughput (FPS) is measured on the test split using single-image batches on the same GPU. Parameters and GFLOPs are profiled at a $1024 \times 1024$ reference input using \texttt{ultralytics} for YOLO models and the Hugging Face \texttt{FlopCountAnalysis} utility for DAB-DETR~\cite{liu2022dabdetr}.

\subsubsection{Detection Results}

Table~\ref{tab:benchmark} presents detection results for all six architectures retrained on the curated dataset, alongside Exsclaim~\cite{schwenker2023exsclaim}, the prior published method in materials science (specifically for microscopic data). Exsclaim~\cite{schwenker2023exsclaim} is trained on classes A-H at an image resolution of $640 \times 640$, which represents its optimal operating condition; its metrics are therefore not directly comparable to our models evaluated across the full sub-panel label set, but serve as an indicative reference for prior work. Each architecture was fine-tuned on a fixed train/val/test split (seed 42) and evaluated on the held-out 281-image test set.

\begin{table}[H]
\centering
\caption{Detection results on the curated test split (281 images). We evaluated at a confidence threshold (conf) of 0.55 (for precision, recall, and F1) and NMS IoU threshold of 0.45, with all models run at an input resolution of 1024$\times$1024. AP\textsubscript{f} / AP\textsubscript{c} denote Average Precision averaged over IoU thresholds 0.50-0.95 (AP@50:95) on frequent and common classes, respectively. $^\star$Exsclaim~\cite{schwenker2023exsclaim} is evaluated under its optimal reported condition. Bold indicates the best result among retrained models.}
\label{tab:benchmark}
\resizebox{\linewidth}{!}{%
\begin{tabular}{lrrrrrrrrrrr}
\toprule
Model & Params & GFLOPs & FPS
      & P & R & F1
      & mAP$_{50}$ & mAP$_{75}$ & mAP$_{50:95}$
      & APf & APc \\
      & (M) & & (img/s) & & & & & & \\
\midrule
Exsclaim$^\star$~\cite{schwenker2023exsclaim}
    & 56.84 & 194.45 & 14.38
    & 0.8286 & 0.6823 & 0.7453
    & 0.7902 & 0.7757 & 0.7598
    & 0.7862 & 0.7157 \\
\midrule
DAB-DETR~\cite{liu2022dabdetr}
    & 43.65 & 64.17 & 9.18
    & 0.9525 & 0.8150 & 0.8784
    & 0.6569 & 0.6500 & 0.6122
    & 0.7538 & 0.5110 \\
\midrule
YOLO11-m~\cite{yolo11_ultralytics}
    & 20.07 & 87.40 & 11.96
    & 0.9302 & 0.9113 & 0.9206
    & 0.8821 & 0.8581 & 0.8273
    & 0.8576 & 0.8057 \\
\midrule
YOLO10-m~\cite{wang2024yolov10}
    & 16.51 & 82.05 & 13.40
    & 0.9701 & 0.9023 & 0.9349
    & 0.8805 & 0.8718 & 0.8344
    & 0.8580 & 0.8174 \\
\midrule
YOLO9-c~\cite{wang2024yolov9}
    & 25.55 & 132.83 & 13.61
    & 0.9371 & 0.9188 & 0.9279
    & 0.8911 & 0.8862 & 0.8504
    & 0.8717 & 0.8352 \\
\midrule
YOLO8-m~\cite{yolov8_ultralytics}
    & 25.87 & 101.29 & 14.31
    & 0.9390 & 0.9256 & 0.9322
    & 0.9028 & 0.8935 & 0.8566
    & 0.8678 & 0.8486 \\
\midrule
YOLO12-m~\cite{tian2025yolov12}
    & 20.15 & 86.82 & 13.76
    & 0.9601 & \textbf{0.9406} & \textbf{0.9502}
    & \textbf{0.9227} & \textbf{0.8989} & \textbf{0.8686}
    & \textbf{0.8955} & \textbf{0.8495} \\
\bottomrule
\end{tabular}}
\end{table}
Among the retrained models, YOLO12-m~\cite{tian2025yolov12} offers the best accuracy while remaining compact (20.15\,M parameters, 86.82\,GFLOPs). Its lead is clearest on mAP$_{50}$, where it reaches 0.9227 against 0.9028 for the next-best YOLO8-m~\cite{yolov8_ultralytics}; on F1 the top YOLO variants are bunched together (0.9279 to 0.9502). For that, we use mAP$_{50}$ rather than F1 as the deciding metric for model selection. Every retrained YOLO model clears Exsclaim~\cite{schwenker2023exsclaim} on mAP$_{50}$ by a wide margin, from 9.0 percentage points for YOLO10-m~\cite{wang2024yolov10} to 13.3\,pp for YOLO12-m~\cite{tian2025yolov12}. We read this as evidence that the gain comes from domain-specific retraining rather than the choice of architecture.

DAB-DETR~\cite{liu2022dabdetr} is the clear outlier. Despite its NMS-free bipartite-matching design~\cite{carion2020end}, it trails every YOLO variant on mAP$_{50}$ (0.6569) while carrying the most parameters (43.65\,M) and running slowest (9.18\,img/s), behaviour expected from transformer detectors when training data is scarce~\cite{dosovitskiy2020image}. With only 2,249 training images, it does not generalise well: precision is high (0.9525), but recall is markedly lower (0.8150), so the panels it does predict are usually correct, though it misses a sizeable share altogether. The YOLO variants present cleaner trade-offs. YOLO10-m~\cite{wang2024yolov10} has the highest precision (0.9701) but gives up recall (0.9023), which is the wrong way round for compound-figure parsing, since a missed panel costs us more than a spurious box. YOLO8-m~\cite{yolov8_ultralytics} pairs solid mAP$_{50}$ (0.9028) with the fastest inference in the group (14.31\,img/s), making it the better pick when speed matters most. YOLO12-m~\cite{tian2025yolov12} offers the best balance across accuracy, size, and speed, and we adopt it as the sub-panel detector for the annotation pipeline.

\subsection{Caption, Category, and Summary Generation}
\label{sec:annotation}

After the detector localises the sub-panels, each panel enters the second stage of the pipeline, where a large language model (LLM) produces its structured annotation. The model sees only text at this point: the original figure caption and the in-text reference sentence. The panel image itself is never passed to the LLM. Specifically, for each atomic image unit (sub-panel/panel), the pipeline produces three outputs: (i) a \textbf{caption}, grounded from the relevant portion of the original figure caption and the in-text reference sentence, (ii) a \textbf{visualisation category and sub-category}, classifying the image type according to a predefined materials science taxonomy; and (iii) a concise \textbf{summary} of the image content and its scientific significance.

\subsubsection{Taxonomy Design}
To enable systematic classification of materials science figures, we define a two-level taxonomy comprising 19 broad visualisation categories, each associated with a set of domain-specific subtypes. The taxonomy was first drafted from subject-area knowledge of materials characterisation techniques, refined in consultation with materials science PhD-level researchers. The categories span the full range of experimental and computational representations encountered in materials research, including \textit{Microscopy} (e.g., SEM, TEM, STEM, AFM), \textit{Diffraction} (e.g., XRD Pattern, EBSD Map, SAED), \textit{Spectroscopy} (e.g., XPS, Raman, FTIR, EDX), \textit{Mechanical Test} (e.g., Stress-Strain Curve, Hardness Map), \textit{Electrochemistry} (e.g., Cyclic Voltammogram, Nyquist Plot), and \textit{Simulation} (e.g., DFT Result, MD Snapshot, Phase-Field Simulation), among others. Categories such as \textit{Schematic/Diagram}, \textit{Photograph}, and \textit{Generic Plot} are also included to capture non-quantitative and illustrative content. The full taxonomy is provided in the Appendix~\ref{sec:appendix_taxonomy}. Both the category and subtype fields are specified as closed vocabularies in the prompt, with \textit{other} reserved as a fallback only when no defined subtype is applicable. 

\subsubsection{Annotation Generation}
Structured annotations are generated by prompting an LLM with the original figure caption and the in-text reference sentence extracted during source data collection. We give the model the full two-level taxonomy directly in the prompt, written out as a readable block that lists each broad category with its valid subtypes, and we instruct it to pick both fields only from these options. We enforce this at the API level with a JSON schema: the visualisation category is a strict enumeration, every response field is required, and no extra properties are allowed, so the output is always well-formed and needs no post-hoc parsing. The subtype is constrained by the prompt rather than the schema, since the chosen category already narrows its valid values. We run the model at a low temperature of 0.1 to keep the outputs deterministic and faithful to the text, and disable chain-of-thought reasoning to keep throughput high.

The prompt tells the model to ground each sub-caption strictly in the supplied caption and reference sentence and not to speculate beyond what the text says. Reference sentences are applied selectively: a group-level reference that supports the whole caption is shared across all panels, whereas a panel-specific reference is scoped to its own sub-panel. To avoid repetitive phrasing, we instruct the model to vary how it opens each sentence across panels of the same figure. Summaries are capped at 40 to 60 words and drawn only from the panel's own sub-caption and the reference sentences tied to it; they must not refer to the figure or panel labels, so the text stays on the scientific content. The full prompt template shown in Figure~\ref{fig:prompt_template}, Appendix~\ref{sec:appendix_prompt}.

The output for each sub-panel is a structured record containing the panel identifier, visualisation category, visualisation subtype, sub-caption, and summary, which together form the final annotation triplet alongside the cropped sub-panel image.
\subsection{Benchmarking}
\label{sec:benchmarking}
To evaluate the quality of model-generated annotations in a systematic and reproducible manner, we developed a custom web-based annotation tool using Flask. We benchmarked six large language models: Gemini 3.5 Flash, DeepSeek V4 Flash, Gemini 3.1 Flash Lite, Mistral Large 3, GPT-5.4 Mini, and GPT-5.4 Nano, on two aspects of annotation quality: (i) the accuracy of visualisation category and sub-category classifications, and (ii) the quality of generated sub-captions and summaries. All six models were selected for their low inference cost, making them suitable for large-scale annotation pipelines. The annotation tool presents the evaluator with the original compound figure alongside the original caption, reference sentences, and the model-generated category, subcategory, sub-caption, and summary for each model, side by side. Because the evaluator inspects the actual image, the ground truth reflects what each panel really shows, which lets us measure whether the text-derived classifications and generated annotations match the image content. The source code for the annotation tool is available in the accompanying code repository.

\subsubsection{Category Classification}
\begin{figure}[H]
    \centering
    \includegraphics[width=0.99\linewidth]{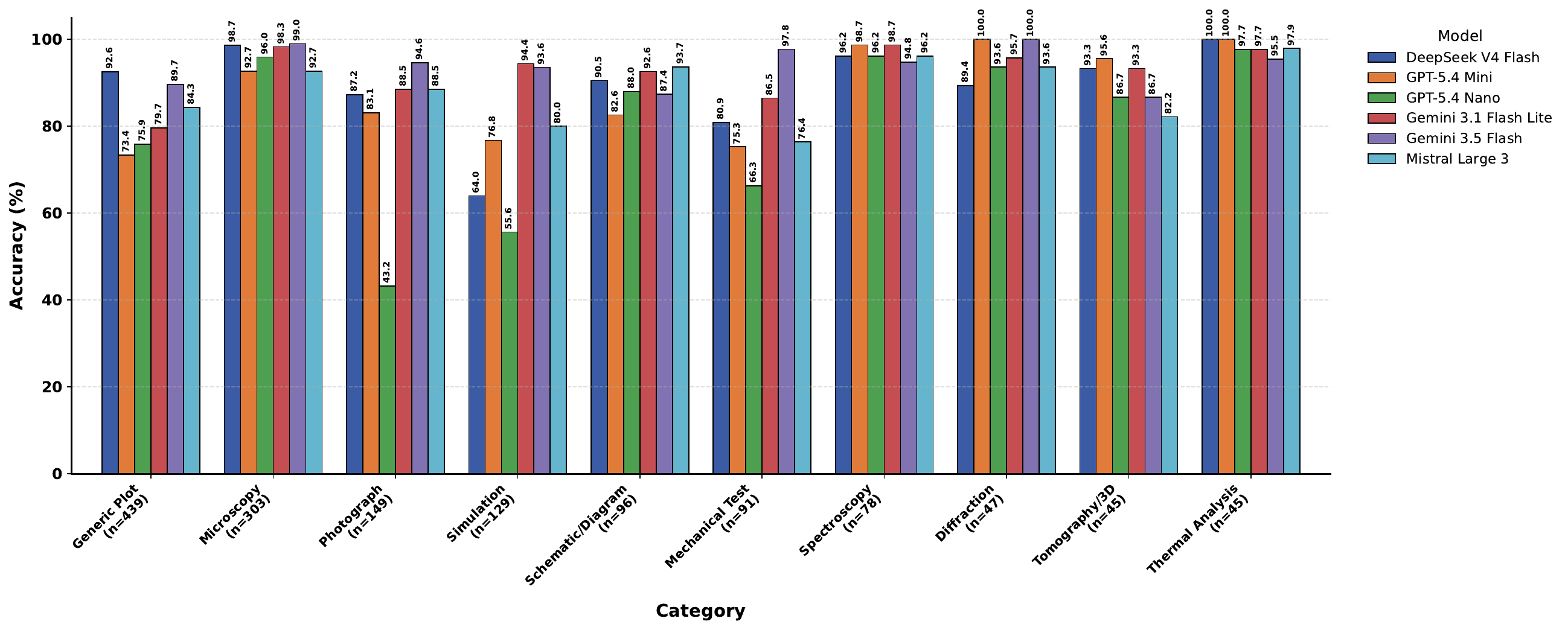}
    \caption{Per-category classification accuracy (\%) across six LLMs on a 350 compound figure evaluation subset. Only categories with more than 20 annotated atomic panels are reported. Sample counts per category are shown on the x-axis.}
    \label{fig:category_acc}
\end{figure}

We benchmarked category classification on a subset of 350 compound figures sampled from the dataset. As caption segmentation itself is model-dependent, the number of atomic panels identified varies slightly across models. For instance, a model may merge two adjacent sub-panels (e.g., panels (d) and (e)) into a single prediction, or omit a panel entirely. The resulting atomic panel counts were: Gemini 3.1 Flash Lite (1512), Gemini 3.5 Flash (1527), GPT-5.4 Mini (1518), GPT-5.4 Nano (1511), DeepSeek V4 Flash (1528), and Mistral Large 3 (1539). Since the evaluator inspects the original compound figure directly within the annotation tool, these segmentation discrepancies are visually identifiable against the source image. Per-category accuracy is reported as the percentage of each model's own predicted panels that were correctly classified relative to the visually verified ground truth. Figure~\ref{fig:category_acc} reports per-category accuracy across all six models; we restrict the comparison to categories with more than 20 annotated atomic panels, below which the per-category estimates become too noisy to interpret.

Gemini 3.5 Flash is the strongest model overall, and its advantage is largest on categories with distinctive visual and terminological signatures: 99.0\% on Microscopy, 97.8\% on Mechanical Test, and 95.5\% on Thermal Analysis. Gemini 3.1 Flash Lite is close behind and holds up well across the board, leading all six models on Simulation (94.4\%) and scoring above 98\% on Spectroscopy and Microscopy.

Photograph is the hardest category for every model. GPT-5.4 Nano is the weakest here at 43.2\%, which is unsurprising given that photographic content is often only identifiable from the image itself, and the caption text provides little to disambiguate it. Simulation is the other persistent weak spot, where GPT-5.4 Mini (76.8\%) and GPT-5.4 Nano (55.6\%) fall well short of the Gemini models. DeepSeek V4 Flash is a mixed case: it is competitive on the high-frequency categories (98.7\% on Microscopy, 92.6\% on Generic Plot) but drops to 64.0\% on Simulation, so its weaknesses are concentrated in specific categories rather than spread evenly across the taxonomy.

\subsubsection{Sub-Category Classification}
\begin{figure}[H]
    \centering
    \includegraphics[width=0.99\linewidth]{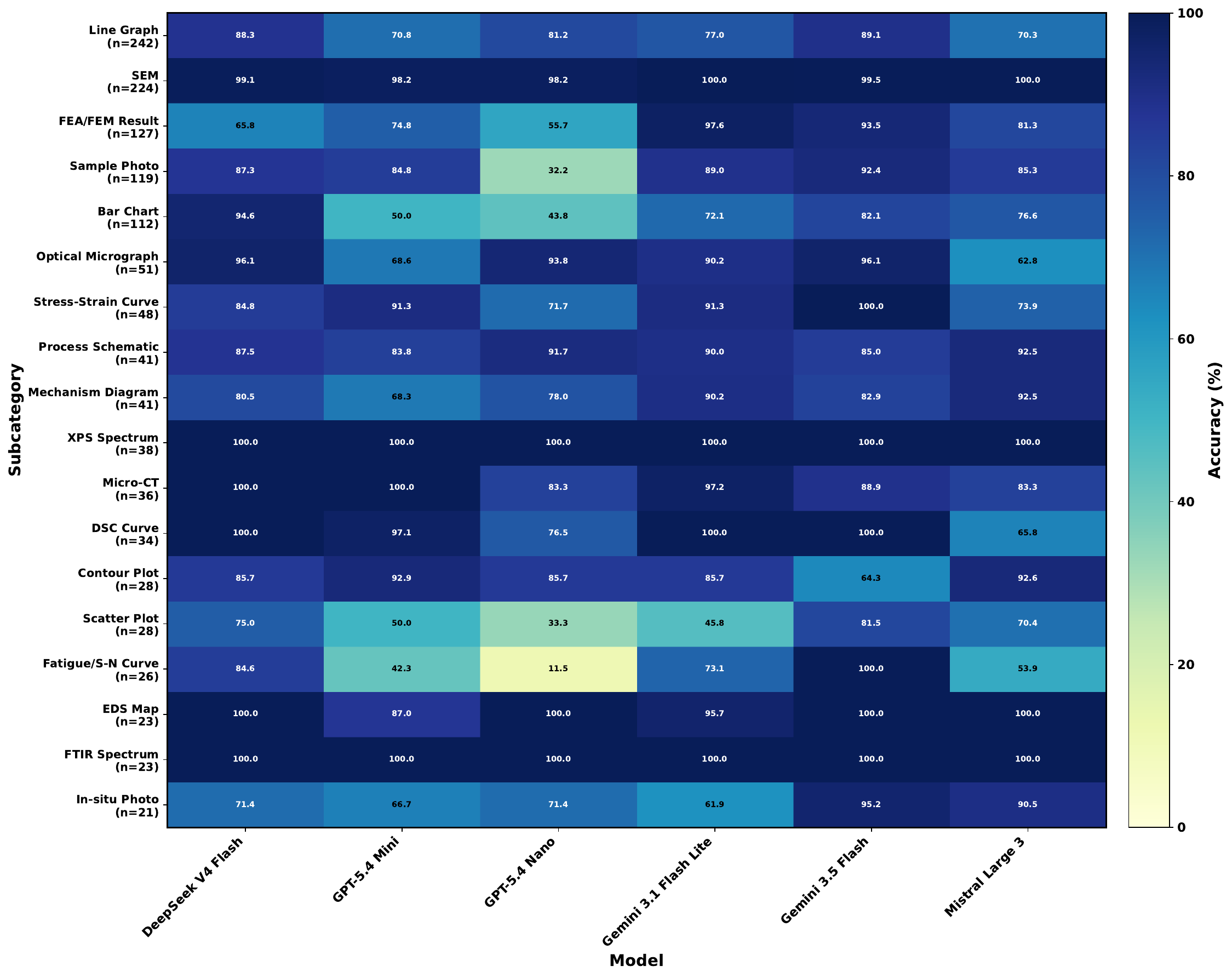}
    \caption{Per-subcategory classification accuracy (\%) across six LLMs shown as a heatmap, evaluated on the same 350-figure subset. Only subcategories with more than 20 annotated atomic panels are reported. Sample counts per subcategory are shown on the y-axis.}
    \label{fig:subcategory_acc}
\end{figure}

We evaluate sub-category classification on the same 350-figure subset, using the same atomic panel counts. Figure~\ref{fig:subcategory_acc} shows per-subcategory accuracy as a heatmap across the six models, again restricted to subcategories with more than 20 annotated panels.

A few subcategories are easy for every model. XPS Spectrum and FTIR Spectrum reach perfect or near-perfect accuracy throughout, which follows from their distinctive spectral signatures. SEM is similar, with all six models having accuracy above 98\%.

The hard cases cluster among the generic plot types. Since all models are conditioned on the same caption and reference sentence, the wide spread in accuracy on a single subtype indicates that the relevant terminology is usually present or inferable in the source text, but the weaker models fail to reliably extract and map it onto the correct taxonomy entry, often defaulting instead to a more generic or frequent subtype. Fatigue/S-N Curve illustrates this most clearly: GPT-5.4 Nano manages only 11.5\% and GPT-5.4 Mini 42.3\%. Gemini 3.5 Flash scores a full 100\%, showing that the distinguishing cues are recoverable from text when the model is capable of using them. Scatter Plot and Bar Chart vary just as widely, with GPT-5.4 Nano at 33.3\% and 43.8\% respectively, so it is the fine-grained plot-type distinctions where the smaller models break down. Sample Photo splits the models the same way: GPT-5.4 Nano falls to 32.2\% while most others stay above 84\%.

DeepSeek V4 Flash is strong and steady here, scoring perfectly on EDS Map, XPS Spectrum, FTIR Spectrum, DSC Curve, and Micro-CT, and holding up on harder subcategories such as FEA/FEM Result (65.8\%). Taking the section as a whole, Gemini 3.5 Flash and DeepSeek V4 Flash are the most reliable for subcategory classification, while GPT-5.4 Nano is the most erratic and the weakest on fine-grained distinctions. Subcategories with small sample counts ($n < 30$) should be treated as indicative only, since their accuracies carry wide uncertainty.

\subsubsection{Sub-Caption and Summary Generation}
\begin{figure}[H]
    \centering
    \includegraphics[width=0.99\linewidth]{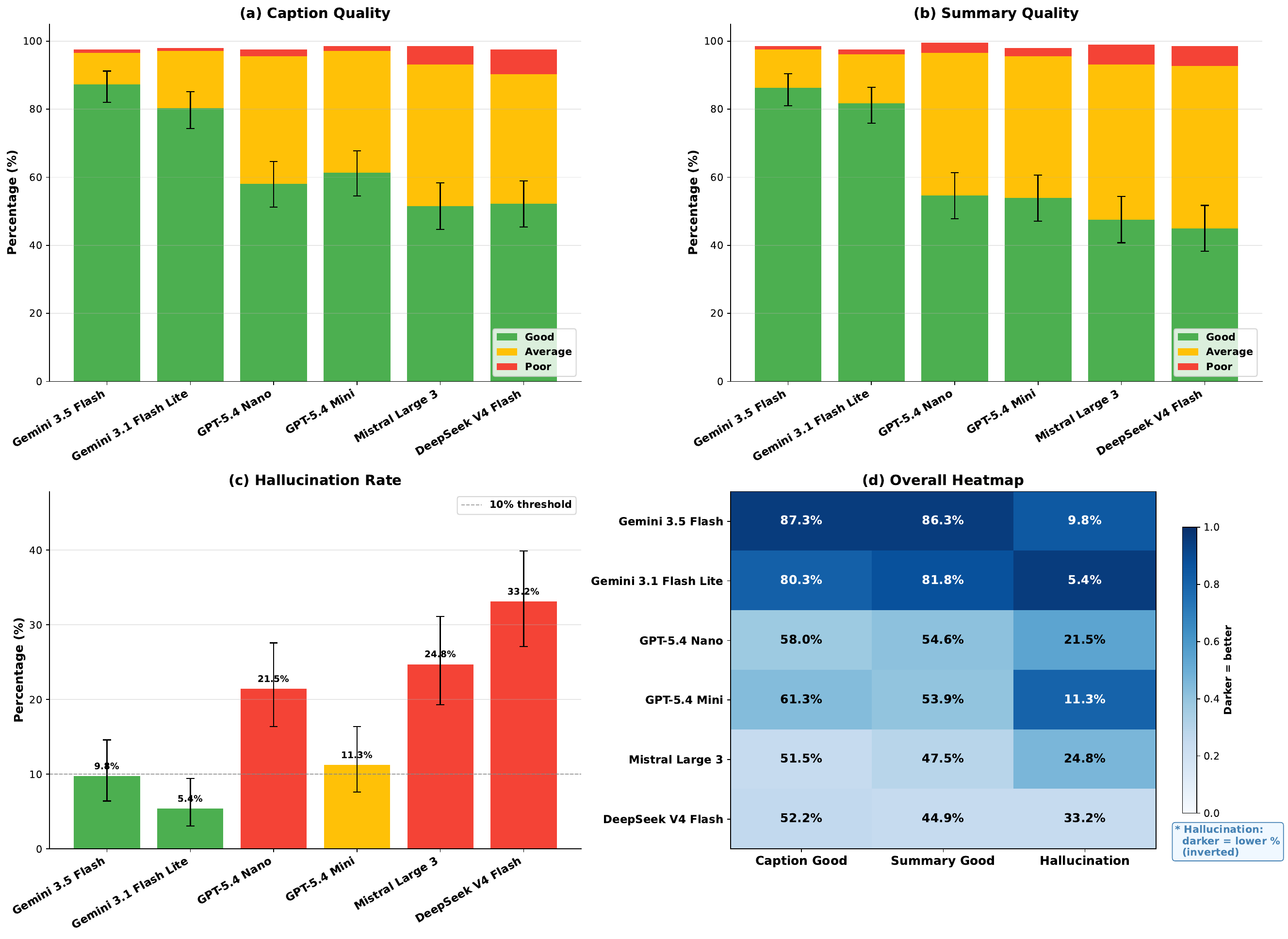}
    \caption{Sub-caption and summary generation quality across six LLMs, evaluated on 40 compound figures (1{,}224 panel-level annotations). (a) Caption quality distribution. (b) Summary quality distribution. (c) Hallucination rate per model, with the 10\% acceptability threshold shown as a dashed line. (d) Heatmap summarising caption, summary, and hallucination rates. Quality labels (good/average/poor) and hallucination were assessed independently, so a panel rated good or average may still have been flagged for hallucination. Error bars and bracketed intervals denote 95\% Wilson~\cite{Wilson01061927} confidence intervals.}
    \label{fig:caption_summary_bench}
\end{figure}

Sub-caption and summary quality were benchmarked on the 40 compound figures in the evaluation set (a subset of 350 figures), yielding 1{,}224 panel-level sub-caption/summary pairs across the six models. Per-model counts were: Gemini 3.5 Flash (205), GPT-5.4 Nano 
(205), DeepSeek V4 Flash (205), GPT-5.4 Mini (204), Gemini 3.1 Flash Lite (203), and Mistral Large 3 (202). Each compound figure was associated with its original caption and in-text reference sentence, and the same six models were used to generate sub-captions and summaries for every atomic sub-panel from this textual information alone. We define a \textit{hallucination} as any statement in a generated sub-caption or summary that introduces information absent from, and not inferable from, the source figure caption and in-text reference sentence. That was deemed as fabricated content with no textual grounding, as distinct from paraphrase or omission. A domain expert assessed each output via the annotation interface, assigning one of three qualitative labels: \textit{good}, \textit{average}, or \textit{poor}. It has been done independently for the sub-caption and summary, and hallucination has been flagged where present. Quality is reported as the percentage of panels in each label, and hallucination as the percentage of panels explicitly flagged, each computed over all evaluated panels for that model. All proportions are reported with 95\% Wilson confidence~\cite{Wilson01061927} intervals, which are appropriate for the sample sizes and low event rates involved. We note that the labels are provided by a single materials science PhD student. Figure~\ref{fig:caption_summary_bench} summarises the results across all six models.
The two Gemini models lead in both quality and reliability. Gemini 3.5 Flash has the highest quality, with 87.3\% [82.1, 91.2] of sub-captions and 86.3\% [81.0, 90.4] of summaries rated good, and a hallucination rate of 9.8\% [6.4, 14.6], just under the 10\% threshold. Gemini 3.1 Flash Lite is close behind, at 80.3\% [74.3, 85.2] of sub-captions and 81.8\% [75.9, 86.5] of summaries rated good; these intervals overlap those of Gemini 3.5 Flash, so the two are statistically indistinguishable on generation quality. Gemini 3.1 Flash Lite also has the lowest hallucination rate we observed, 5.4\% [3.1, 9.4], and is the only model whose entire confidence interval sits below 10\%.

The other four models trail by a wide margin. GPT-5.4 Nano and GPT-5.4 Mini land at caption good-rates of 58.0\% and 61.3\%, and summary good-rates of 54.6\% and 53.9\%, while DeepSeek V4 Flash and Mistral Large 3 
are weaker still (52.2\%/44.9\% and 51.5\%/47.5\% caption/summary, respectively). All four cross the hallucination threshold, from 11.3\% [7.6, 16.3] for GPT-5.4 Mini up to 33.2\% [27.1, 39.9] for DeepSeek V4 Flash. The DeepSeek result is telling: its strong category classification does not carry over to faithful text generation, a reminder that classification accuracy and generation reliability are separate capabilities.

The picture is therefore a clear split between the Gemini family and everything else. Gemini 3.5 Flash and Gemini 3.1 Flash Lite are statistically comparable on caption and summary quality, but Gemini 3.1 Flash Lite has a lower hallucination rate and costs a fraction as much to run (Section~\ref{sec:cost}), so we select it as the annotation model for all downstream sub-caption and summary generation in the final pipeline.
\subsubsection{Inference Infrastructure and Cost}
\label{sec:cost}
We access all models through two cloud platforms: the Gemini family via Google AI Studio~\cite{googleaistudio}, and DeepSeek V4 Flash, Mistral Large 3, GPT-5.4 Mini, and GPT-5.4 Nano via Microsoft Azure AI Foundry~\cite{microsoftazurefoundry}. Table~\ref{tab:model_pricing} lists the per-token pricing for each model at the time of evaluation.

\begin{table}[H]
\centering
\caption{Inference pricing for the six benchmarked models. All costs are in USD per million tokens.}
\label{tab:model_pricing}
\begin{tabular}{lrr}
\toprule
\textbf{Model} & \textbf{Input (\$/1M tokens)} & \textbf{Output (\$/1M tokens)} \\
\midrule
DeepSeek V4 Flash~\cite{deepseek_v4_flash_pricing}      & 0.19 & 0.51 \\
GPT-5.4 Nano~\cite{azure_openai_pricing}           & 0.20 & 1.25 \\
Gemini 3.1 Flash Lite~\cite{google_gemini_pricing}  & 0.25 & 1.50 \\
Mistral Large 3~\cite{mistral_large_3_pricing}        & 0.50 & 1.50 \\
GPT-5.4 Mini~\cite{azure_openai_pricing}           & 0.75 & 4.50 \\
Gemini 3.5 Flash~\cite{google_gemini_pricing}       & 1.50 & 9.00 \\
\bottomrule
\end{tabular}
\end{table}

Models are ordered by increasing input cost. DeepSeek V4 Flash and GPT-5.4 Nano represent the most cost-efficient options at \$0.19 and \$0.20 per million input tokens, respectively, while Gemini 3.5 Flash incurs the highest inference cost at \$1.50 and \$9.00 per million input and output tokens. However, cost alone does not determine model selection. The benchmarking results demonstrate that Gemini 3.5 Flash and Gemini 3.1 Flash Lite substantially outperform the remaining models on sub-caption quality, hallucination rate and category classification. In contrast, the lower-cost models exhibit hallucination rates well above the acceptable threshold. Considering both cost and quality, Gemini 3.1 Flash Lite emerges as the most practical choice for large-scale annotation, offering strong generation quality and the lowest hallucination rate of 5.4\% at a modest cost of \$0.25 per million input tokens. It is therefore selected as the model for all downstream annotation tasks in this pipeline.
\section{The \texttt{MatMMExtract} Package}

The complete pipeline described in this work is implemented as \texttt{matmmextract}, an open-source Python package (Python\,$\geq$\,3.10) available on the Python Package Index (PyPI) and installable via \texttt{pip install matmmextract}. The package is organised into six modules corresponding to the pipeline stages described in this paper. Full documentation and usage examples are available in the accompanying code repository. Two examples are shown using this package in Appendix~\ref{sec:appendix_qualitative_example}.
\section{Dataset}
Table~\ref{tab:dataset_comparison} compares MatSciFig with the existing materials science multimodal datasets. While prior work has produced valuable benchmarks for evaluation~\cite{lai-etal-2025-multimodal, weng2025matqna, mcgrath2026matrix} and knowledge extraction~\cite{liu2026multimodal}, none provides a large-scale, panel-level dataset suitable for training vision-language models across the full breadth of materials science visualisation modalities. EXSCLAIM!~\cite{schwenker2023exsclaim} is the closest in spirit, offering panel-level annotations at scale, but is restricted to electron microscopy and provides keyword-level NLP extractions rather than grounded natural language descriptions. MatSciFig addresses this gap by combining panel-level decomposition, structured visualisation taxonomy classification, and LLM-generated sub-captions and summaries. The result is a training-ready resource several times larger than previous panel-level figure datasets in materials science.

\begin{table}[H]
\centering
\caption{Comparison of existing materials science figure datasets with MatSciFig.}
\label{tab:dataset_comparison}
\resizebox{\textwidth}{!}{%
\begin{tabular}{l p{3.5cm} p{3.5cm} p{4cm} p{1.8cm} p{1.8cm} p{2cm}}
\toprule
\textbf{Dataset} & \textbf{Size} & \textbf{Modality} & \textbf{Annotation Type} & \textbf{Panel-level} & \textbf{Purpose} & \textbf{License} \\
\midrule
EXSCLAIM!~\cite{schwenker2023exsclaim} & 83,504 figures & Electron Microscopy & NLP keyword extraction & Yes & Training & CC-BY \\
\midrule
MatCha~\cite{lai-etal-2025-multimodal} & 1,260 panels/ 2,165 figures & SEM, TEM, XRD, XPS & GPT-4o, expert review; VQA & Yes & Evaluation & CC-BY \\
\midrule
MatQnA~\cite{weng2025matqna} & 4,968 QA pairs & SEM, TEM, XRD, XPS, AFM, TGA, DSC, FTIR, Raman, XAFS & GPT-4.1, human validation; VQA & No & Evaluation & -- \\
\midrule
MATRIX~\cite{mcgrath2026matrix} & 3,056 panels & SEM, XRD, EDS, TGA & LLM-generated, rubric judge & Partial & Evaluation & -- \\
\midrule
MatMech~\cite{liu2026multimodal} & 425,295 figures/ 207,200 mechanisms & SEM, TEM, XRD, Spectroscopy & LLM, expert validation; causal chains & No & Knowledge Base & CC-BY \\
\midrule
\textbf{MatSciFig (Ours)} & \textbf{391,606 panels / 180,571 figures} & \textbf{19 modality categories} & \textbf{LLM-generated; sub-caption, category, summary} & \textbf{Yes} & \textbf{Training} & \textbf{CC-BY-NC} \\
\bottomrule
\end{tabular}}
\end{table}
\subsection{MatSciFig}
We release \textbf{MatSciFig}, a large-scale panel-level dataset of scientific figures drawn from materials science literature. MatSciFig is constructed by applying the full pipeline described in this work to open-access articles from the alloy, composite, and ceramic research domains. Compound figures are decomposed into individual atomic panels, each paired with a panel-specific sub-caption, visualisation category and sub-category, and an extended summary derived from the source paper. The dataset comprises 391,606 panels with full metadata extracted from 180,571 source figures, making it one of the largest figure-level vision-language datasets in the materials science domain.
MatSciFig is derived from articles published under CC-BY and CC-BY-NC licences, both of which require attribution for redistributed derivative works. It is publicly available on the Hugging Face Hub and is released under CC-BY-NC 4.0, the more restrictive of the two source licences, to ensure compliance across the full corpus.

\begin{table}[H]
\centering
\caption{MatSciFig dataset schema.}
\label{tab:dataset_columns}
\begin{tabular}{lll}
\toprule
\textbf{Column} & \textbf{Type} & \textbf{Description} \\
\midrule
\texttt{image}                    & bytes  & Panel crop in original format \\
\texttt{image\_id}                & string & Parent figure identifier \\
\texttt{panel\_suffix}            & string & Panel label (A, B, C\,\ldots\ or single) \\
\texttt{visualization\_category}  & string & High-level category (e.g., Microscopy, Generic Plot) \\
\texttt{visualization\_subtype}   & string & Fine-grained subtype (e.g., SEM, XRD Pattern) \\
\texttt{subcaption}               & string & Panel-level caption derived from the source paper \\
\texttt{summary}                  & string & Extended description of the panel content \\
\bottomrule
\end{tabular}
\end{table}
\subsection{Data Quality and Post-Processing}
\label{sec:data_quality}
Following compound figure detection and LLM-based annotation, the two stages are joined by matching each cropped panel filename to its corresponding entry in the LLM output using the panel label as the key (e.g., \texttt{imgXXXX\_B.jpg} $\rightarrow$ key \texttt{b}), with single-panel figures mapped to the reserved key \texttt{main}. Of the 469,061 detected panels, 391,606 (83.5\%) are successfully matched and retained in the final dataset. The remaining 16.5\% are excluded: 2.7\% due to missing LLM output files and 13.8\% due to panel key mismatches. Most mismatches occur when the detector localises panels that are not explicitly labelled in the caption and therefore absent from the LLM annotation.

Annotation quality is further assessed by checking whether each subtype is a valid member of its assigned category. Since the visualisation category is enforced as a strict enumeration via the JSON schema, all 391,606 panels carry a valid category. The subtype, being prompt-guided rather than schema-enforced, is a valid child of its assigned category in 380,458 panels (97.15\%). A further 2,969 panels (0.76\%) carry the reserved fallback value \textit{other} under a category other than \textit{Other}; the prompt permits \textit{other} whenever no listed subtype applies, so these are treated as intended behaviour rather than as deviations. The remaining 8,179 panels (2.09\%) fall into three classes. In 5,420 panels (1.38\%) the subtype field holds a category name, most often \textit{Generic Plot} (5,381 panels), which the model occasionally assigns at the wrong level of the taxonomy. In 2,546 panels (0.65\%) the subtype is a valid taxonomy entry but not a child of the assigned category, such as \textit{CALPHAD Diagram} under \textit{Simulation} (386 panels) or \textit{EBSD IPF Map} under \textit{Diffraction} (208). The remaining 213 panels (0.05\%), spanning 49 unique strings, carry values absent from the taxonomy altogether; the most frequent are \textit{MD Result} (35), a near miss paraphrase of the defined \textit{MD Snapshot} and \textit{MD Trajectory} subtypes, followed by \textit{Pie Chart} (23), \textit{Infrared Thermography} (15), and \textit{Mott-Schottky Plot} (14). All panels are retained in the dataset, and a per-panel validity flag is released alongside the metadata so that users requiring strictly conformant labels can filter.

To confirm that detector-assigned panel letters map to the correct physical panel, we manually audited 937 matched panels, verifying each subcaption's specific details against its assigned crop. Of these, 915 (97.7\%) were correctly aligned, 21 (2.2\%) were ambiguous (e.g., a partially cropped panel letter or merged sub-labels), and only 1 (0.1\%) was a confirmed misalignment, indicating that the letter-based join introduces negligible label noise.
\subsection{Dataset Statistics}
\begin{figure}[H]
    \centering
    \includegraphics[width=0.85\linewidth]{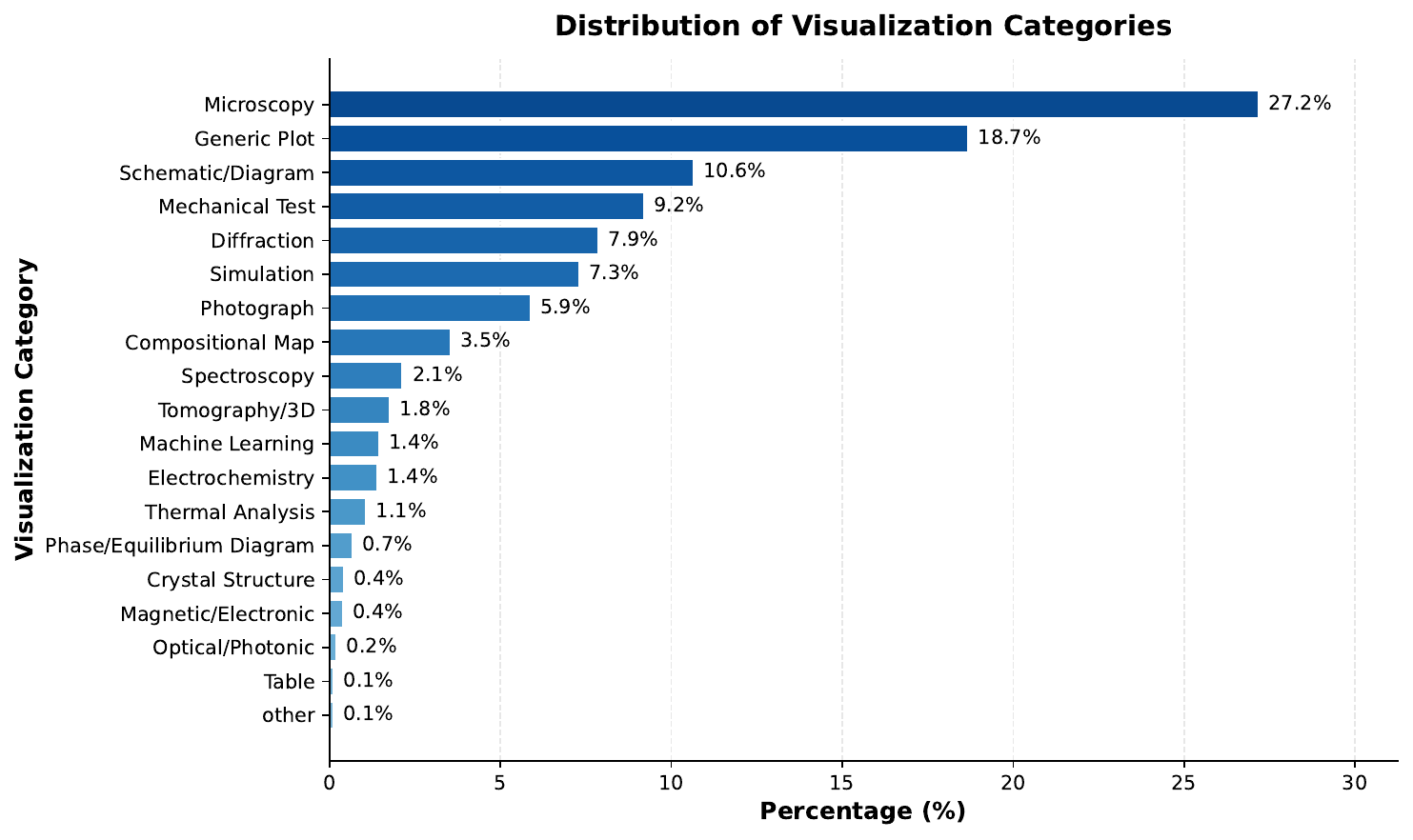}
    \caption{Distribution of panels across visualisation categories in MatSciFig.}
    \label{fig:category_dist}
\end{figure}

Figure~\ref{fig:category_dist} shows how panels are distributed across visualisation categories. Microscopy dominates at 27.2\% of all panels, which fits the central role of SEM and TEM imaging in characterising materials microstructure. Generic Plot (18.7\%) and Schematic/Diagram (10.6\%) come next, reflecting how often quantitative plots and process illustrations appear in materials science papers. Mechanical Test (9.2\%) and Diffraction (7.9\%) follow. Smaller but scientifically important categories are also present, including Tomography/3D (1.8\%), Electrochemistry (1.4\%), and Crystal Structure (0.4\%). 

\begin{figure}[H]
    \centering
    \includegraphics[width=0.99\linewidth]{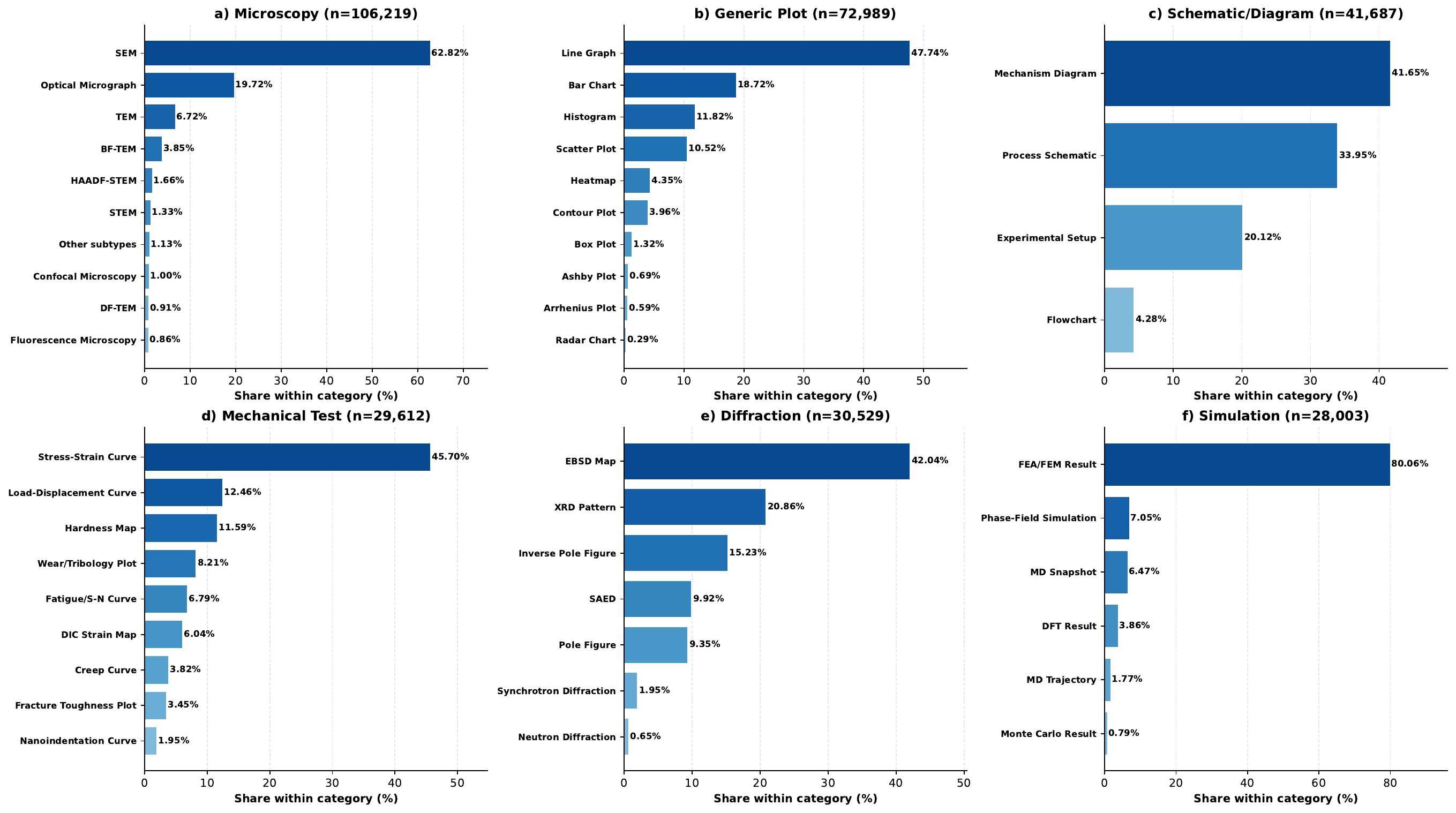}
    \caption{Sub-category distribution within the six most populated visualisation categories in MatSciFig: (a) Microscopy, (b) Generic Plot, (c) Schematic/Diagram, (d) Mechanical Test, (e) Diffraction, and (f) Simulation. Sample counts for each category are shown in the subplot titles.}
    \label{fig:subcategory_dist}
\end{figure}

Figure~\ref{fig:subcategory_dist} breaks down the subtypes within the six most populated categories, restricted to panels whose subtype is a valid child of the assigned category (Section~\ref{sec:data_quality}). Microscopy is led by SEM at 62.8\%, then Optical Micrograph (19.7\%) and TEM (6.7\%), which mirrors how heavily microstructure characterisation relies on SEM and optical imaging due to their easier availability and ease of sample preparation. In Generic Plot, Line Graph (47.7\%) and Bar Chart (18.7\%) are the most common subtypes. Mechanical Test is dominated by Stress-Strain Curve (45.7\%), ahead of Load-Displacement Curve (12.5\%) and Hardness Map (11.6\%). Diffraction is led by EBSD Map (42.0\%) and XRD Pattern (20.9\%), and Simulation even more so by FEA/FEM Result (80.1\%), in line with the computational emphasis of structural materials research.

\begin{figure}[H]
    \centering
    \includegraphics[width=0.85\linewidth]{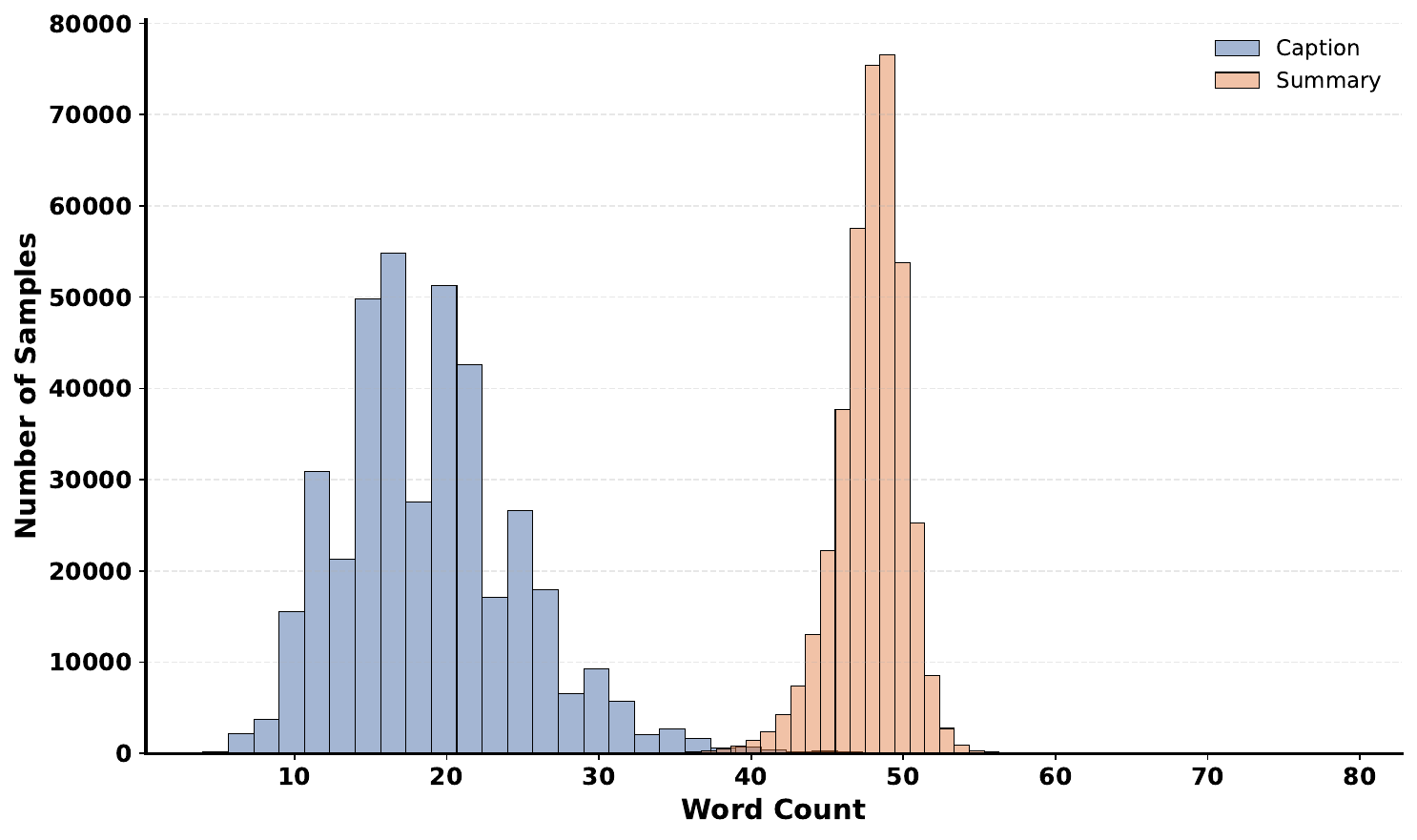}
    \caption{Word count distribution of sub-captions and summaries across the 391,606 panels in MatSciFig.}
    \label{fig:wordcount_dist}
\end{figure}

Figure~\ref{fig:wordcount_dist} shows the word-count distributions of sub-captions and summaries across the dataset. Sub-captions spread out, centred around 15 to 20 words but running from under 10 to over 35, which tracks the natural variation in caption length across the source papers. Summaries are much tighter, concentrated between 45 and 55 words and mostly at 48-50, a direct result of the 40-60 word limit we impose during generation. The two behave differently by design: sub-captions stay close to the brevity of the original captions, whereas summaries give each panel a longer and more uniform description.

\section{Retrieval Experiment}

\subsection{Model Architecture}
To demonstrate the utility of MatSciFig for vision-language learning, we train a dual-encoder retrieval model that maps panel images and their corresponding text descriptions into a shared embedding space. The model consists of two independent branches, each followed by an identical projection head.

The image branch uses a CLIP ViT-B/32 vision encoder~\cite{radford2021learning}, extracting the CLS representation as a 768-dimensional feature vector. The text branch uses MatSciBERT~\cite{gupta2022matscibert}, a BERT-base model pre-trained on materials science literature, extracting the CLS token from the final hidden state as a 768-dimensional feature vector. Each panel is paired with its summary as the text input; where a summary is unavailable, the subcaption is used as a fallback. Both branches are followed by a bias-free linear layer, LayerNorm, and L2-normalisation, mapping each modality into a shared 512-dimensional embedding space.

\subsection{Training Objective}
The model is trained with symmetric InfoNCE loss~\cite{oord2018representation}, consistent with the CLIP training paradigm~\cite{radford2021learning}. Given a batch of $B$ image-text pairs, the similarity matrix $S \in \mathbb{R}^{B \times B}$ is computed as:

\begin{equation}
    S = \frac{\mathbf{e}_{\text{img}} \cdot \mathbf{e}_{\text{txt}}^\top}{\tau}
\end{equation}
\noindent where $\mathbf{e}_{\text{img}}$ and $\mathbf{e}_{\text{txt}}$ are the L2-normalised image and text embedding matrices, and $\tau$ is a learnable temperature parameter initialised to 0.07. The training loss is the average of cross-entropy losses computed over both directions:

\begin{equation}
    \mathcal{L} = \frac{\mathcal{L}_{i \rightarrow t} + \mathcal{L}_{t \rightarrow i}}{2}
\end{equation}
\noindent where the diagonal entries of $S$ correspond to true pairs and the off-diagonal entries serve as in-batch negatives. By the end of training, $\tau$ converged to 0.026, indicating that the model learned to produce well-separated embeddings with a sharper similarity distribution than the initialisation.

\subsection{Training Setup}

\subsubsection{Data Split}
The dataset is split at the figure level to prevent panel leakage across splits. Panels from the same source figure are assigned to the same subset. Splits are stratified by dominant visualisation subtype and follow an 80/10/10 panel-count ratio. The resulting splits contain 312,971 training panels, 39,320 validation panels, and 39,315 test panels. Subtypes with fewer than 10 total panels are assigned entirely to the training set to avoid empty evaluation buckets. All splits are generated with a fixed random seed of 42 for reproducibility.
\subsubsection{Optimiser and Schedule}
The model is trained for 20 epochs using AdamW with $\varepsilon = 1 \times 10^{-6}$ and weight decay of 0.01, excluding bias parameters and LayerNorm weights. Three separate learning rate groups are used to account for the different levels of domain adaptation required by each component: the CLIP vision encoder, which undergoes a larger domain shift from general web images to materials science imagery, is fine-tuned at $3 \times 10^{-5}$; the MatSciBERT text encoder, which is already domain-adapted, is fine-tuned at the lower rate of $2 \times 10^{-5}$; and the projection heads and temperature parameter, trained from scratch, use a higher learning rate of $1 \times 10^{-4}$. A cosine decay schedule with 10\% linear warmup is applied across all parameter groups. The model is trained with a per-step batch size of 128 and gradient accumulation over 2 steps, giving an effective batch size of 256. Gradients are clipped to a maximum norm of 1.0 and training is conducted in bfloat16 precision. Table~\ref{tab:optimiser} summarises the full optimiser configuration.
\begin{table}[H]
\centering
\caption{Optimiser configuration for dual-encoder training.}
\label{tab:optimiser}
\begin{tabular}{ll}
\toprule
\textbf{Parameter} & \textbf{Value} \\
\midrule
Optimiser              & AdamW \\
$\varepsilon$          & $1 \times 10^{-6}$ \\
Weight decay           & 0.01 (bias / LayerNorm excluded) \\
LR - vision encoder  & $3 \times 10^{-5}$ \\
LR - text encoder    & $2 \times 10^{-5}$ \\
LR - projection + $\tau$ & $1 \times 10^{-4}$ \\
Batch size (effective) & 256 \\
Schedule               & Cosine decay + 10\% linear warmup \\
Gradient accumulation   & 2  \\
Gradient clip          & 1.0 \\
Epochs                 & 20 \\
Total steps            & $\approx$24,450 \\
Precision              & bfloat16 \\
\bottomrule
\end{tabular}
\end{table}
\subsubsection{Convergence}
\begin{figure}[H]
    \centering
    \includegraphics[width=0.85\linewidth]{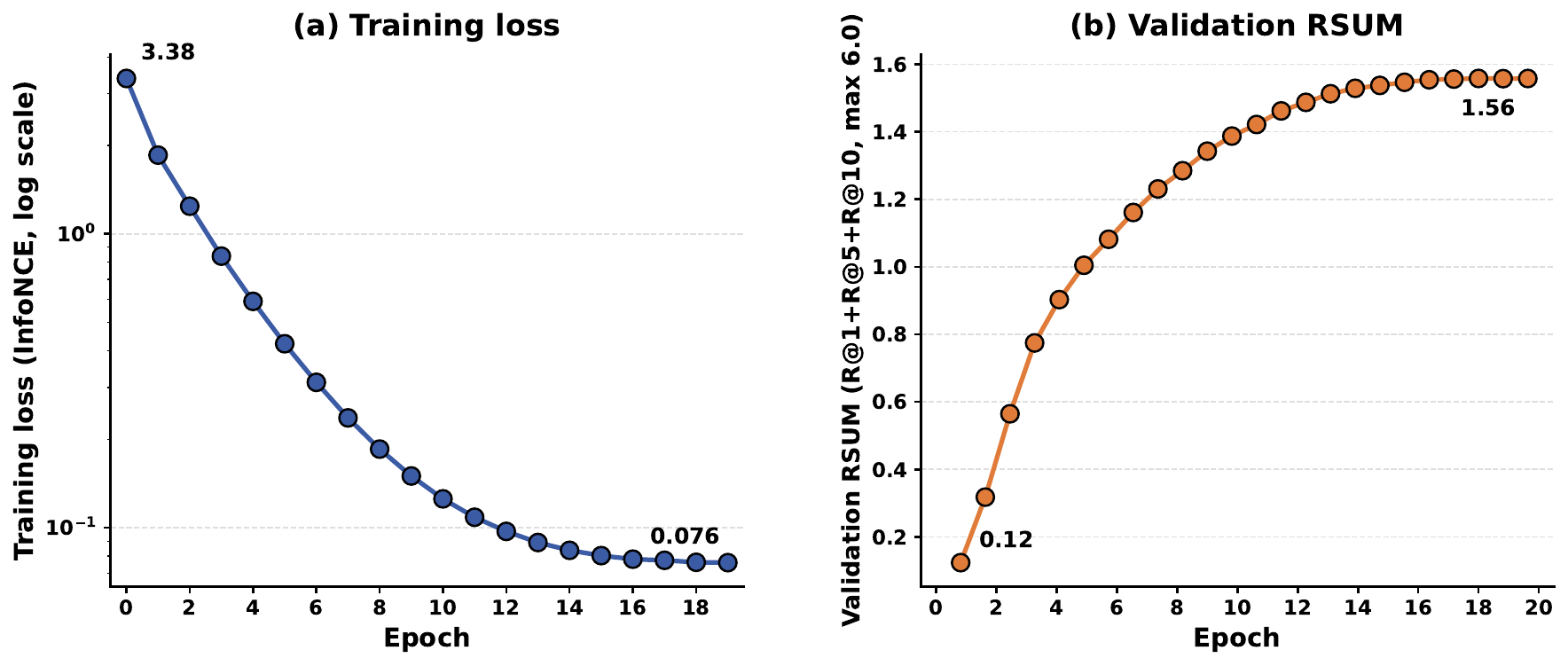}
    \caption{Training convergence over 20 epochs. (a) Average InfoNCE training loss per epoch (log scale). (b) Validation RSUM (R@1+R@5+R@10 summed across both retrieval directions) at each evaluation checkpoint.}
    \label{fig:convergence}
\end{figure}
The model converges steadily over 20 epochs, with the average training loss decreasing from 3.38 at epoch 0 to 0.076 by epoch 19 (Figure~\ref{fig:convergence}a), a roughly 44-fold reduction with no signs of instability or divergence. Validation RSUM, defined as the sum of R@1, R@5, and R@10 across both retrieval directions, improves from 0.12 at the first evaluation checkpoint (step 1{,}000) to 1.56 at the final epoch, with the most rapid gains occurring in the first two epochs before the curve flattens (Figure~\ref{fig:convergence}b).
\subsection{Evaluation Protocol}
All models are evaluated on the held-out test split of 39,315 panels. Panel embeddings are computed once at inference and indexed using FAISS~\cite{8733051}, which performs exact inner-product search on L2-normalised embeddings, equivalent to cosine similarity retrieval. Evaluation is conducted in both directions: image-to-text (i$\rightarrow$t), where an image embedding is used to query the text gallery, and text-to-image (t$\rightarrow$i), where a text embedding queries the image gallery. We report Recall at K (R@K), the percentage of queries for which the correct match appears within the top-$K$ retrieved results, for $K \in \{1, 5, 10, 50, 100\}$, along with Mean Reciprocal Rank (MRR) computed at depth 1000. The model checkpoint selected by the best validation RSUM, evaluated every 1,000 training steps, is used for all reported results. Per category Recall values are reported only for the categories with at least 200 test panels, as fewer samples are unreliable.

\subsection{Results}
\subsubsection{Overall Retrieval Performance}
Table~\ref{tab:retrieval_overall} reports overall retrieval performance against two baselines: a random retrieval baseline and zero-shot CLIP, which uses the pretrained CLIP ViT-B/32 without any fine-tuning. Zero-shot CLIP achieves modest performance, with R@1 of 2.4\% and 1.7\% in the i$\rightarrow$t and t$\rightarrow$i directions, respectively, reflecting the substantial domain gap between general web images and materials science panels. Fine-tuning on MatSciFig produces consistent and substantial gains across all metrics: R@1 improves to 11.2\% (i$\rightarrow$t) and 9.1\% (t$\rightarrow$i), representing improvements of $4.7\times$ and $5.4\times$ over zero-shot CLIP, respectively. R@10 reaches 40.5\% and 38.1\%, and R@100 reaches 71.3\% and 70.1\%, demonstrating that the fine-tuned model places the correct panel within the top 100 retrieved results for approximately 70\% of queries. MRR improves from 0.043 to 0.206 in the i$\rightarrow$t direction, indicating that correct matches are consistently ranked near the top of the retrieved list.

\begin{table}[H]
\centering
\caption{Overall cross-modal retrieval performance on 39,315 test panels. R@K values in \%. Best result per row in \textbf{bold}.}
\label{tab:retrieval_overall}
\begin{tabular}{lccccc}
\toprule
Metric & Random & \multicolumn{2}{c}{Zero-shot CLIP} & \multicolumn{2}{c}{\textbf{Ours (Fine-tuned)}} \\
\cmidrule(lr){3-4} \cmidrule(lr){5-6}
       &        & i$\rightarrow$t & t$\rightarrow$i & \textbf{i$\rightarrow$t} & \textbf{t$\rightarrow$i} \\
\midrule
R@1    & 0.0   & 2.4   & 1.7   & \textbf{11.2} & 9.1  \\
R@5    & 0.0   & 5.7   & 3.9   & \textbf{30.1} & 27.1 \\
R@10   & 0.0   & 7.8   & 5.5   & \textbf{40.5} & 38.1 \\
R@50   & 0.1   & 14.8  & 11.1  & \textbf{63.2} & 62.1 \\
R@100  & 0.3   & 19.3  & 14.9  & \textbf{71.3} & 70.1 \\
MRR    & 0.000 & 0.043 & 0.031 & \textbf{0.206} & 0.182 \\
\bottomrule
\end{tabular}
\end{table}
\subsubsection{Per Category Retrieval Performance}
Table~\ref{tab:retrieval_category} presents per-category retrieval results. Fine-tuning consistently improves performance across all categories in both retrieval directions, with no category regressing to zero-shot CLIP performance. Performance varies considerably across categories, largely reflecting the degree of visual and textual distinctiveness of each class.

Categories with highly distinctive visual signatures achieve the strongest results. Phase/Equilibrium Diagram achieves the highest R@1 of 23.1\% (i$\rightarrow$t) and an MRR of 0.359, consistent with its visually and textually distinctive content. Tomography/3D (R@1 = 19.0\%, MRR = 0.314), Machine Learning (R@1 = 19.9\%, MRR = 0.320), and Electrochemistry (R@1 = 18.1\%, MRR = 0.303) also perform strongly, benefiting from the specificity of their associated visual representations and descriptions.

The most challenging category remains Microscopy, which accounts for the largest share of the test set ($n = 10{,}624$) and achieves the lowest R@1 of 10.0\% (i$\rightarrow$t). To understand these failure patterns, we manually inspected retrieval errors. For Microscopy and Diffraction, the model correctly identifies the visualization subtype (e.g., retrieving another SEM image or EBSD map) but fails to distinguish the specific instance, since captions largely describe imaging modality and generic microstructural vocabulary (e.g., ``reveals the microstructure,'' ``grain refinement'') rather than material- or condition-specific detail, and the underlying panels are themselves visually similar. Diffraction (R@1 = 12.0\%) and Schematic/Diagram (R@1 = 14.2\%) present similar challenges, with large intra-category visual variability making retrieval inherently harder. Generic Plot performs better than Microscopy despite being large ($n = 7{,}305$), achieving an R@1 of 16.7\%; the likely reason is that plot types such as line graphs and bar charts have distinctive structural cues that align well with how they are described in the text.

\begin{table*}[t]
\centering
\caption{Per-category retrieval performance for both directions (i$\rightarrow$t: image queries text; t$\rightarrow$i: text queries image). Each category is evaluated against its own within-category gallery. R@K values in \%. Categories sorted by test-set size. Categories with fewer than 200 test panels are excluded.}
\label{tab:retrieval_category}
\resizebox{\textwidth}{!}{
\begin{tabular}{lrcccccccccc}
\toprule
& & \multicolumn{4}{c}{Zero-shot CLIP} & \multicolumn{6}{c}{\textbf{Ours (Fine-tuned)}} \\
\cmidrule(lr){3-6} \cmidrule(lr){7-12}
& & \multicolumn{2}{c}{i$\rightarrow$t} & \multicolumn{2}{c}{t$\rightarrow$i} & \multicolumn{3}{c}{\textbf{i$\rightarrow$t}} & \multicolumn{3}{c}{\textbf{t$\rightarrow$i}} \\
\cmidrule(lr){3-4} \cmidrule(lr){5-6} \cmidrule(lr){7-9} \cmidrule(lr){10-12}
Category & $n$ & R@1 & R@10 & R@1 & R@10 & \textbf{R@1} & \textbf{R@10} & \textbf{MRR} & \textbf{R@1} & \textbf{R@10} & \textbf{MRR} \\
\midrule
Microscopy              & 10,624 & 1.1  & 4.0  & 0.5 & 2.7  & \textbf{10.0} & \textbf{36.8} & \textbf{0.187} & \textbf{8.3}  & \textbf{34.5} & \textbf{0.167} \\
Generic Plot            & 7,305  & 5.7  & 16.5 & 3.9 & 12.4 & \textbf{16.7} & \textbf{52.1} & \textbf{0.282} & \textbf{13.9} & \textbf{49.1} & \textbf{0.252} \\
Schematic/Diagram       & 4,205  & 6.7  & 19.2 & 4.1 & 11.8 & \textbf{14.2} & \textbf{46.3} & \textbf{0.244} & \textbf{11.0} & \textbf{44.1} & \textbf{0.215} \\
Mechanical Test         & 3,637  & 4.2  & 13.5 & 2.4 & 8.8  & \textbf{17.5} & \textbf{53.3} & \textbf{0.291} & \textbf{13.6} & \textbf{48.5} & \textbf{0.247} \\
Diffraction             & 3,090  & 1.5  & 6.8  & 1.2 & 5.3  & \textbf{12.0} & \textbf{42.2} & \textbf{0.218} & \textbf{9.5}  & \textbf{37.7} & \textbf{0.188} \\
Simulation              & 2,844  & 4.0  & 14.1 & 2.8 & 8.9  & \textbf{16.4} & \textbf{50.7} & \textbf{0.278} & \textbf{13.9} & \textbf{49.3} & \textbf{0.250} \\
Photograph              & 2,321  & 6.2  & 18.6 & 4.0 & 14.3 & \textbf{14.8} & \textbf{49.4} & \textbf{0.256} & \textbf{12.4} & \textbf{48.0} & \textbf{0.236} \\
Compositional Map       & 1,405  & 1.6  & 5.4  & 0.7 & 4.4  & \textbf{15.5} & \textbf{51.6} & \textbf{0.270} & \textbf{12.8} & \textbf{47.5} & \textbf{0.236} \\
Spectroscopy            & 849    & 3.7  & 15.9 & 3.5 & 13.3 & \textbf{16.8} & \textbf{54.8} & \textbf{0.294} & \textbf{12.7} & \textbf{49.7} & \textbf{0.244} \\
Tomography/3D           & 736    & 5.2  & 14.9 & 4.1 & 13.7 & \textbf{19.0} & \textbf{57.2} & \textbf{0.314} & \textbf{15.1} & \textbf{53.1} & \textbf{0.267} \\
Machine Learning        & 589    & 12.7 & 33.1 & 9.8 & 30.7 & \textbf{19.9} & \textbf{57.4} & \textbf{0.320} & \textbf{16.6} & \textbf{50.3} & \textbf{0.275} \\
Electrochemistry        & 554    & 4.3  & 20.4 & 5.4 & 17.3 & \textbf{18.1} & \textbf{55.6} & \textbf{0.303} & \textbf{11.9} & \textbf{46.4} & \textbf{0.223} \\
Thermal Analysis        & 398    & 6.3  & 20.1 & 5.3 & 16.1 & \textbf{17.8} & \textbf{57.5} & \textbf{0.312} & \textbf{14.8} & \textbf{52.3} & \textbf{0.262} \\
Phase/Equilibrium Diagram & 260  & 8.1  & 30.0 & 6.2 & 22.3 & \textbf{23.1} & \textbf{61.5} & \textbf{0.359} & \textbf{18.8} & \textbf{53.5} & \textbf{0.302} \\
\bottomrule
\end{tabular}
}
\end{table*}

\section{Limitations and Future Work}

MatSciFig has several limitations. The dataset currently covers only open-access articles from Elsevier and Springer, which may bias it toward particular domains and research groups; adding more publishers would widen its coverage. The structured annotations (sub-captions, visualisation categories, and summaries) are generated solely from text, without direct visual grounding. While the benchmarking results demonstrate that this approach produces high-quality annotations for the majority of panels, it is inherently limited by the completeness and clarity of the original figure captions and reference sentences. Panels with ambiguous or insufficiently descriptive captions may receive lower-fidelity annotations, and the hallucination rates observed in lower-cost models further underscore the sensitivity of text-only generation to caption quality.

The current taxonomy, while comprehensive, was designed to cover the most common visualisation types encountered in alloy, composite, and ceramic literature. It may not fully capture the visual diversity across other materials' sub-disciplines within the materials domain, where distinct figure types may be prevalent. Extending and refining the taxonomy in collaboration with domain experts across a wider range of sub-fields is an important direction for future work. The dual-encoder architecture with CLIP ViT-B/32 and MatSciBERT represents a straightforward starting point rather than a state-of-the-art system. Future work should explore larger vision-language models, cross-attention fusion architectures, and retrieval-augmented generation approaches that leverage the rich structured annotations available in MatSciFig. The broader implication of this work is that the same figure-decomposition and structured-annotation strategy that has proven effective for building large-scale, high-quality multimodal resources in other scientific domains can be extended to materials science with comparable success.

\section{Conclusion}

We have presented MatMMExtract, an end-to-end open-source pipeline for constructing panel-level multimodal datasets from materials science literature, and MatSciFig, the large-scale dataset it produces. By combining structured XML parsing from Elsevier and Springer publishers, domain-adaptive compound figure detection using a manually annotated materials science dataset (MaterialScope), and LLM-based structured annotation guided by a two-level visualisation taxonomy, the pipeline transforms raw scientific figures into richly annotated image-text pairs at scale.

MatSciFig comprises 391,606 annotated panels extracted from 180,571 source figures, each paired with a grounded sub-caption, visualisation category and sub-category, and a concise summary, making it one of the largest panel-level multimodal datasets in the materials science domain to date. Benchmarking across six large language models demonstrates that Gemini 3.1 Flash Lite achieves the best balance of annotation quality and cost, with a hallucination rate of only 5.4\%, and is selected for all downstream annotation tasks. The compound figure detector, YOLO12-m fine-tuned on MaterialScope, achieves an mAP$_{50}$ of 0.9227, representing a large improvement over the prior biomedical-trained baseline. Cross-modal retrieval experiments confirm that fine-tuning on MatSciFig produces substantial gains over zero-shot CLIP across all evaluated visualisation categories, validating the dataset's utility for vision-language research in the scientific domain.

We release all components of this work, the MatMMExtract pipeline as a Python package installable through pip, the MaterialScope annotation dataset, the MatSciFig dataset, and the retrieval baseline, as open-source resources to support the development of AI systems capable of understanding and reasoning over the rich visual content of materials science literature. A natural next step is to build on MatSciFig toward richer, reasoning-oriented supervision, moving beyond panel-level captioning to grounded scientific reasoning over figures, mirroring recent progress in domains that have pursued this direction ahead of materials science.

\section*{Code and Data Availability}
The MatSciFig dataset is available at \url{https://huggingface.co/datasets/CMEG-IITR/MatSciFig} and the MaterialScope annotation dataset at \url{https://huggingface.co/datasets/CMEG-IITR/MaterialScope}. The MatMMExtract pipeline is available as a Python package at \url{https://github.com/CMEG-IITR/matmmextract} and installable via \texttt{pip install matmmextract}. All training, benchmarking, and evaluation code is available at \url{https://github.com/CMEG-IITR/cmpfig}.
\section*{Acknowledgements}
The authors would like to thank the India AI Mission for the student fellowship grant. The authors acknowledge that Anthropic's Claude Sonnet 4.6 was used exclusively to improve the grammar, clarity, and readability of this manuscript. All scientific concepts, data analyses, and interpretations were developed entirely by the authors. The authors have carefully reviewed the final manuscript and confirm that the content is technically accurate.

\medskip

\noindent

\medskip

\bibliographystyle{elsarticle-num} 
\bibliography{references}
\appendix
\section{FigEx-Trained Baseline: Domain Gap Analysis}
\label{sec:appendix_figex_baseline}

Table~\ref{tab:figex_val} reports detection performance for all six architectures trained on FigEx~\cite{song-etal-2025-figex} annotations (6,454 train / 717 val biomedical images) for 50 epochs, prior to any domain-adaptive retraining. And Table~\ref{tab:figex_baseline} evaluated on the curated MaterialScope test split. This baseline quantifies the domain gap between biomedical and materials science figures discussed in Section~\ref{sec:compound_figure_detection}, and motivates the domain-adaptive retraining described in Section~\ref{sec:domain_adaptive_training}.
\begin{table}[H]
\centering
\caption{Detection performance of models trained on FigEx~\cite{song-etal-2025-figex} (biomedical) annotations under the same protocol as Table~\ref{tab:benchmark}. APf, APc, and APr denote Average Precision (AP@50:95) on frequent, common, and rare sub-panel classes, respectively.}
\label{tab:figex_val}
\resizebox{\linewidth}{!}{%
\begin{tabular}{lrrrrrrrrr}
\toprule
Model & mAP$_{50}$ & mAP$_{75}$ & mAP$_{50:95}$ & Precision & Recall & F1 & APf & APc & APr \\
\midrule
YOLO12-m~\cite{tian2025yolov12}   & 0.8355 & 0.8286 & 0.7781 & 0.8996 & 0.8461 & 0.8720 & 0.8940 & 0.8486 & 0.5786 \\
\midrule
DAB-DETR~\cite{liu2022dabdetr}    & 0.8021 & 0.7871 & 0.7146 & 0.9658 & 0.9180 & 0.9413 & 0.8624 & 0.7843 & 0.4747 \\
\midrule
YOLO11-m~\cite{yolo11_ultralytics} & 0.8626 & 0.8553 & 0.7999 & 0.8579 & 0.8743 & 0.8660 & 0.8958 & 0.8609 & 0.6331 \\
\midrule
YOLO10-m~\cite{wang2024yolov10}   & 0.8098 & 0.8024 & 0.7530 & 0.8608 & 0.8233 & 0.8416 & 0.8911 & 0.8513 & 0.5052 \\
\midrule
YOLO9-c~\cite{wang2024yolov9}     & 0.9027 & 0.8796 & 0.8371 & 0.9331 & 0.9066 & 0.9197 & 0.8924 & 0.8581 & 0.7509 \\
\midrule
YOLO8-m~\cite{yolov8_ultralytics} & 0.9008 & 0.8925 & 0.8319 & 0.9548 & 0.9085 & 0.9311 & 0.8898 & 0.8580 & 0.7388 \\
\bottomrule
\end{tabular}}
\end{table}

\begin{table}[H]
\centering
\caption{Detection performance of models trained on FigEx~\cite{song-etal-2025-figex} (biomedical) annotations, prior to domain-adaptive training, evaluated on the curated MaterialScope test split under the same protocol as Table~\ref{tab:benchmark}. APf, APc, and APr denote Average Precision (AP@50:95) on frequent, common, and rare sub-panel classes, respectively.}
\label{tab:figex_baseline}
\resizebox{\linewidth}{!}{%
\begin{tabular}{lrrrrrrrrr}
\toprule
Model & mAP$_{50}$ & mAP$_{75}$ & mAP$_{50:95}$ & Precision & Recall & F1 & APf & APc & APr \\
\midrule
YOLO12-m~\cite{tian2025yolov12}   & 0.5438 & 0.5385 & 0.5106 & 0.7353 & 0.5717 & 0.6433 & 0.7127 & 0.4113 & 0.4673 \\
\midrule
DAB-DETR~\cite{liu2022dabdetr}    & 0.5519 & 0.5382 & 0.4806 & 0.8700 & 0.6245 & 0.7271 & 0.7286 & 0.4538 & 0.2754 \\
\midrule
YOLO11-m~\cite{yolo11_ultralytics} & 0.5062 & 0.5028 & 0.4661 & 0.6193 & 0.5356 & 0.5744 & 0.7012 & 0.3913 & 0.3508 \\
\midrule
YOLO10-m~\cite{wang2024yolov10}   & 0.4866 & 0.4820 & 0.4473 & 0.6937 & 0.5131 & 0.5899 & 0.7200 & 0.3954 & 0.2575 \\
\midrule
YOLO9-c~\cite{wang2024yolov9}     & 0.4487 & 0.4357 & 0.4118 & 0.6112 & 0.4666 & 0.5292 & 0.7097 & 0.3904 & 0.1483 \\
\midrule
YOLO8-m~\cite{yolov8_ultralytics} & 0.3834 & 0.3730 & 0.3490 & 0.6068 & 0.4077 & 0.4877 & 0.6783 & 0.2688 & 0.1478 \\
\bottomrule
\end{tabular}}
\end{table}

\section{Visualisation Taxonomy}
\label{sec:appendix_taxonomy}

Table~\ref{tab:taxonomy} presents the full two-level visualisation taxonomy used for classifying sub-panel images in the structured annotation pipeline. Each broad category is associated with a set of domain-specific subtypes that cover the full range of experimental, computational, and illustrative representations encountered in the materials science literature.

\begin{table}[H]
\centering
\caption{Full two-level visualisation taxonomy used for sub-panel classification. The \textit{other} category serves as a fallback when no defined subtype is applicable.}
\label{tab:taxonomy}
\resizebox{\linewidth}{!}{%
\begin{tabular}{ll}
\toprule
\textbf{Category} & \textbf{Subtypes} \\
\midrule
Microscopy & SEM, TEM, STEM, HAADF-STEM, BF-TEM, DF-TEM, Optical Micrograph, \\
           & Confocal Microscopy, AFM, Fluorescence Microscopy, Live/Dead Staining \\
\midrule
Diffraction & XRD Pattern, SAED, EBSD Map, Pole Figure, Inverse Pole Figure, \\
            & Neutron Diffraction, Synchrotron Diffraction \\
\midrule
Spectroscopy & XPS Spectrum, Raman Spectrum, FTIR Spectrum, EDX Spectrum, \\
             & EELS Spectrum, NMR Spectrum, Mass Spectrum, UV-Vis Spectrum, \\
             & Photoluminescence Spectrum, XANES Spectrum, EXAFS Spectrum \\
\midrule
Thermal Analysis & DSC Curve, TGA Curve, DMA Curve, TMA Curve \\
\midrule
Phase/Equilibrium Diagram & Binary Phase Diagram, Ternary Phase Diagram, TTT Diagram, \\
                          & CCT Diagram, CALPHAD Diagram, Pourbaix Diagram \\
\midrule
Mechanical Test & Stress-Strain Curve, Load-Displacement Curve, Nanoindentation Curve, \\
                & Hardness Map, Fatigue/S-N Curve, Creep Curve, Fracture Toughness Plot, \\
                & DIC Strain Map, Wear/Tribology Plot \\
\midrule
Electrochemistry & Cyclic Voltammogram, Charge-Discharge Curve, Capacity Retention Plot, \\
                 & Coulombic Efficiency Plot, Nyquist Plot, Bode Plot, Tafel Plot, \\
                 & Polarization Curve, GITT/PITT Curve, Rate Capability Plot \\
\midrule
Magnetic/Electronic & M-H Hysteresis Loop, M-T Curve, ZFC/FC Curve, I-V Curve, \\
                    & C-V Curve, Band Structure, Density of States, Hall Effect Plot \\
\midrule
Optical/Photonic & Absorbance Spectrum, Transmittance Spectrum, Reflectance Spectrum, \\
                 & EQE/IQE Plot, J-V Curve, Ellipsometry Plot, Refractive Index Plot \\
\midrule
Tomography/3D & APT Reconstruction, Micro-CT, FIB-SEM Tomography, 3D Reconstruction \\
\midrule
Compositional Map & EDS Map, WDS Map, EBSD IPF Map, Elemental Distribution Map \\
\midrule
Simulation & DFT Result, MD Snapshot, MD Trajectory, Phase-Field Simulation, \\
           & FEA/FEM Result, Monte Carlo Result \\
\midrule
Machine Learning & Parity Plot, Confusion Matrix, ROC Curve, Learning Curve, \\
                 & Feature Importance Plot, SHAP Plot, t-SNE/UMAP/PCA Plot \\
\midrule
Crystal Structure & Unit Cell, Atomic Model, Supercell \\
\midrule
Generic Plot & Bar Chart, Scatter Plot, Line Graph, Box Plot, Contour Plot, \\
             & Heatmap, Radar Chart, Ashby Plot, Arrhenius Plot, Histogram \\
\midrule
Schematic/Diagram & Process Schematic, Flowchart, Mechanism Diagram, Experimental Setup \\
\midrule
Photograph & Sample Photo, Equipment Photo, In-situ Photo \\
\midrule
Table & Data Table \\
\midrule
Other & other \\
\bottomrule
\end{tabular}}
\end{table}

\section{Prompt Template}
\label{sec:appendix_prompt}

\begin{figure}[H]
    \centering
    \includegraphics[width=0.99\linewidth]{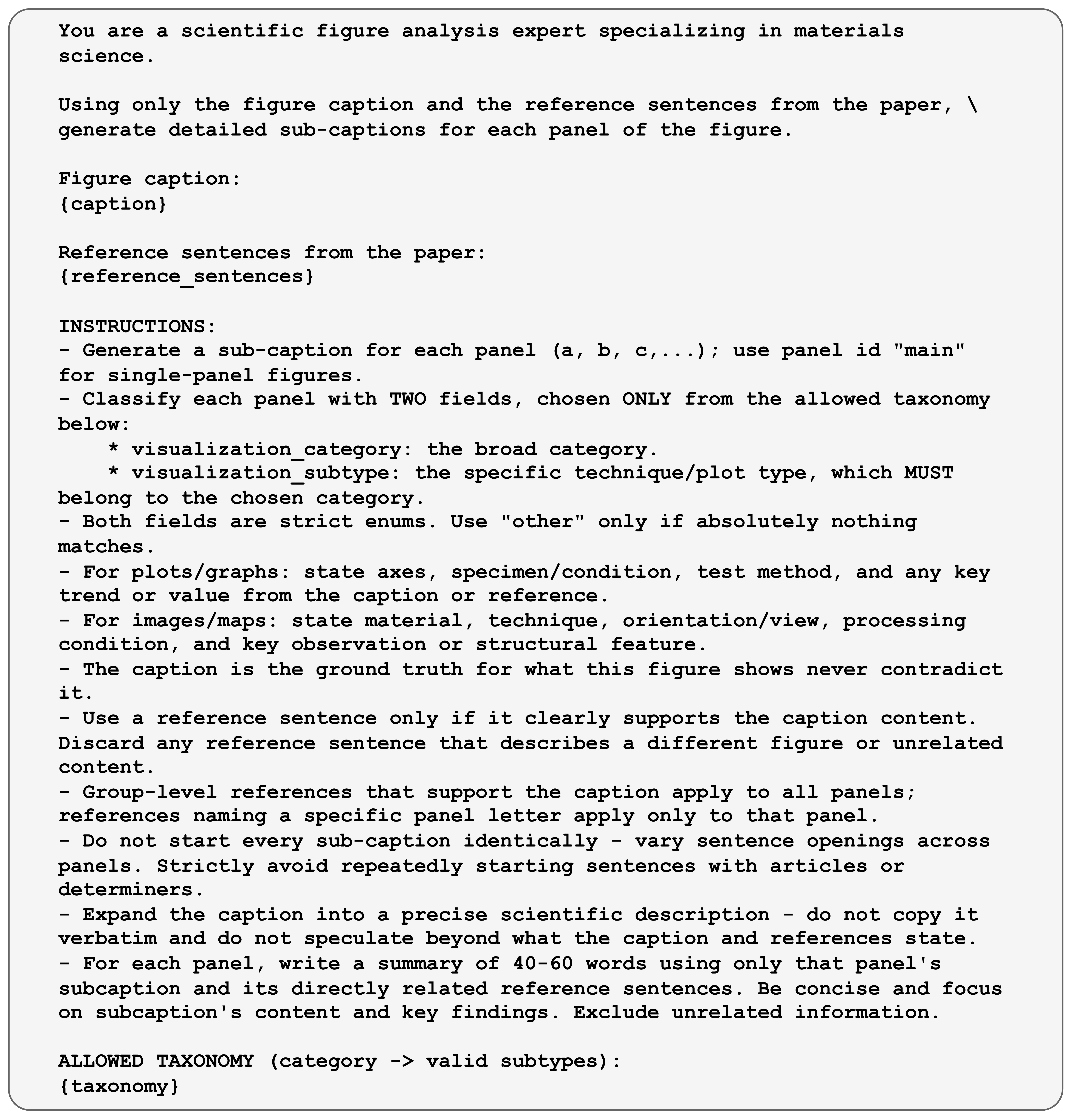}
    \caption{Complete prompt template used for structured annotation generation. The taxonomy block is injected at runtime with the full two-level category and subtype listing.}
    \label{fig:prompt_template}
\end{figure}
\section{Qualitative Pipeline Example}
\label{sec:appendix_qualitative_example}

\begin{figure}[H]
    \centering
    \includegraphics[width=0.99\linewidth]{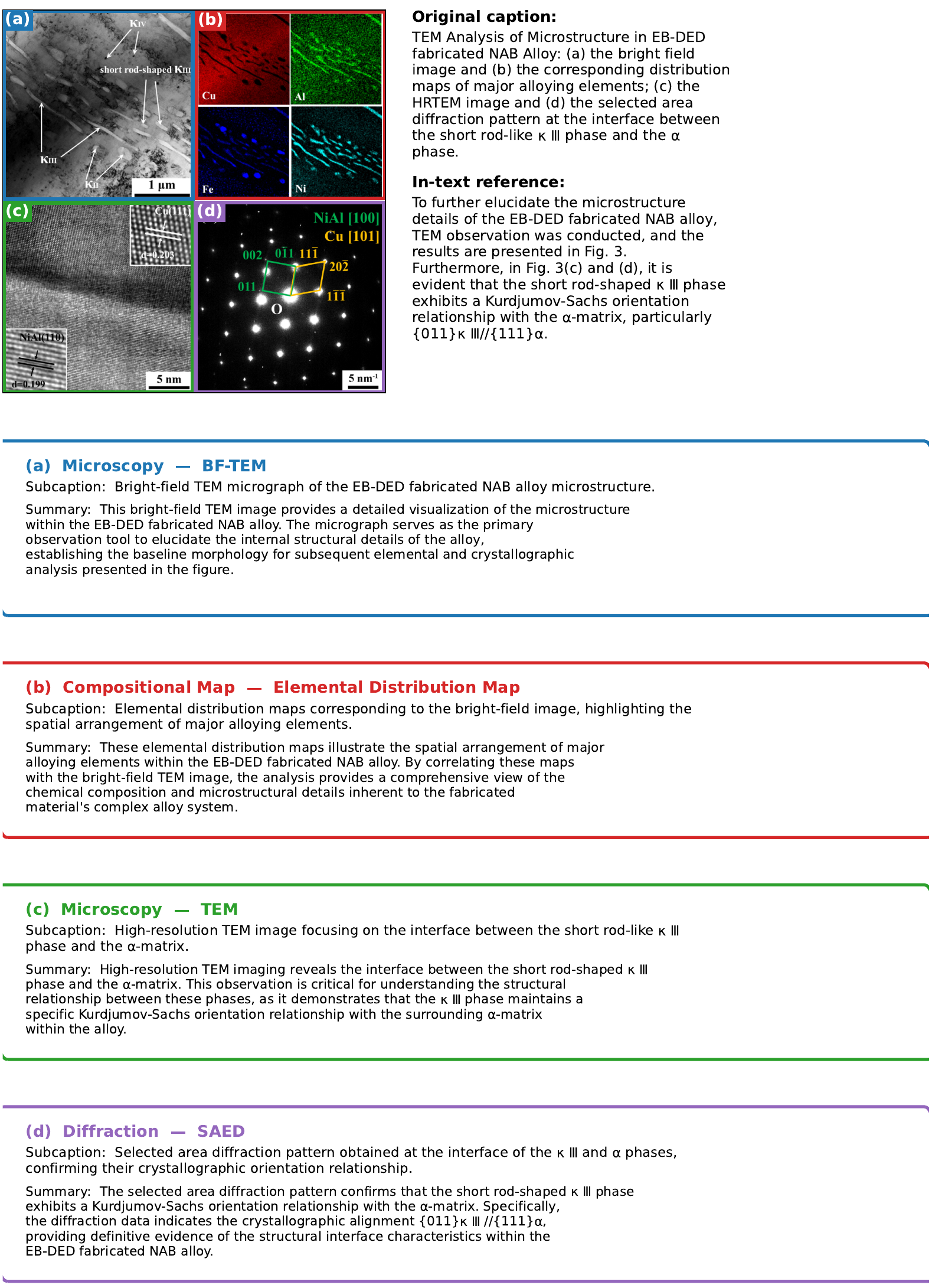}
    \caption{Example of panel detection with original caption/reference (top), followed by per-panel category, subtype, subcaption, and summary annotations generated by the pipeline (bottom) using \textbf{MatMMExtract}}
    \label{fig:appendix_qualitative_example}
\end{figure}

\begin{figure}[H]
    \centering
    \includegraphics[width=0.99\linewidth]{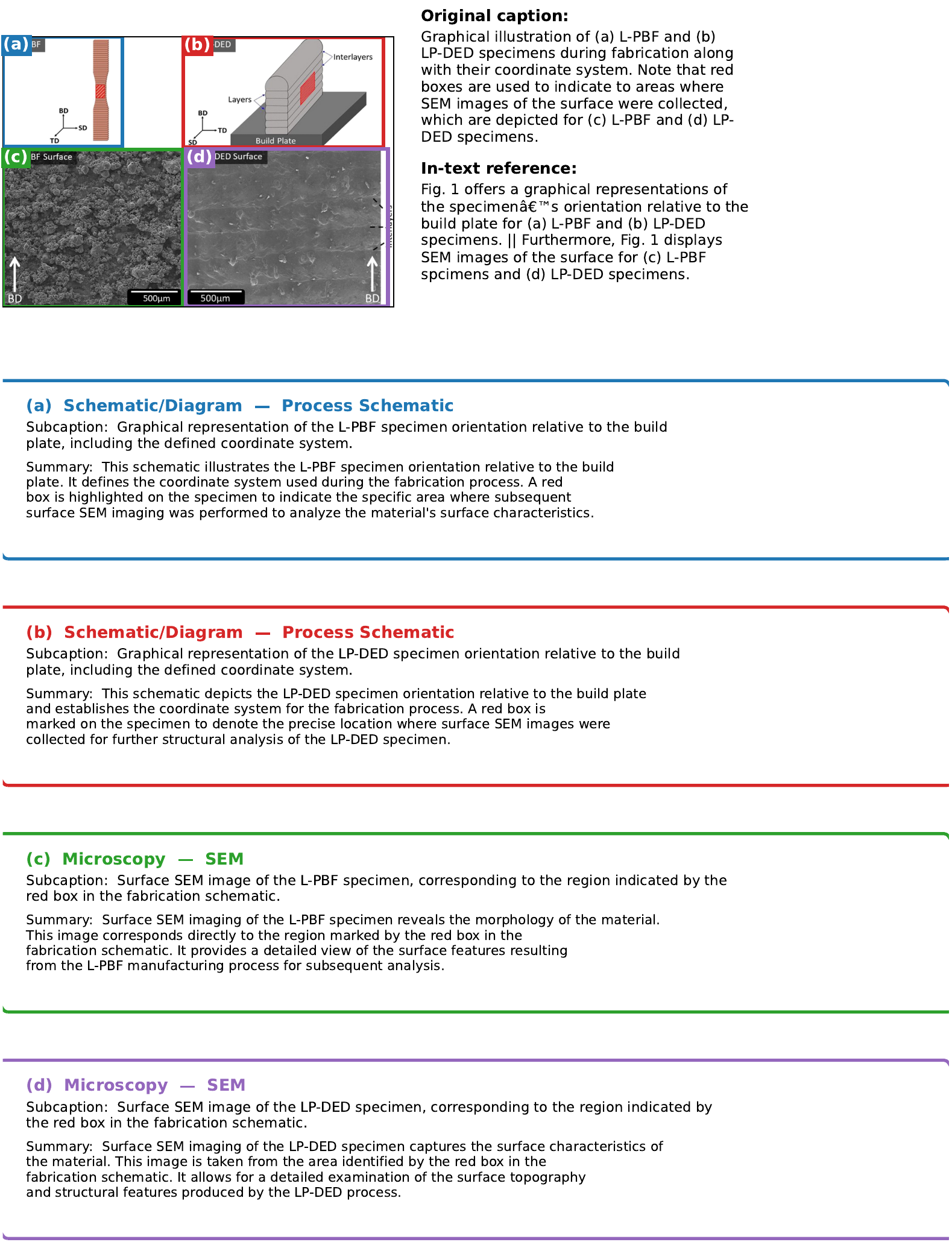}
    \caption{Example of panel detection with original caption/reference (top), followed by per-panel category, subtype, subcaption, and summary annotations generated by the pipeline (bottom) using \textbf{MatMMExtract}}
    \label{fig:appendix_qualitative_example_01}
\end{figure}

Figure~\ref{fig:appendix_qualitative_example} and Figure~\ref{fig:appendix_qualitative_example_01} show an example of the MatMMExtract annotation stage applied to compound figures, illustrating how the schema fields described in Table~\ref{tab:dataset_columns} are populated in practice.


\end{document}